\documentclass[10pt,twocolumn,letterpaper]{article}

\usepackage{cvpr}
\usepackage{times}
\usepackage{epsfig}
\usepackage{graphicx}
\usepackage{amsmath}
\usepackage{amssymb}
\usepackage{breqn}
\usepackage{multirow}

\usepackage{caption}
\DeclareMathOperator{\E}{\mathbb{E}}

\usepackage[pagebackref=true,breaklinks=true,letterpaper=true,colorlinks,bookmarks=false]{hyperref}

\cvprfinalcopy 

\def\cvprPaperID{1902} 

\ifcvprfinal\fi
\begin{document}

\title{I\textsuperscript{n}2I : Unsupervised Multi-Image-to-Image Translation\\ Using Generative Adversarial Networks}

\author{Pramuditha Perera,  Mahdi Abavisani, and Vishal M. Patel\\
Rutgers University, Department of Electrical and Computer Engineering\\
94 Brett Road, Piscataway, NJ 08854, USA\\
{\tt\small  pramuditha.perera@rutgers.edu, mahdi.abavisani@rutgers.edu, vishal.m.patel@rutgers.edu}
}

\setlength\abovedisplayskip{0pt}
\setlength\belowdisplayskip{0pt}

\maketitle
 \begin{abstract}
In unsupervised image-to-image translation, the goal is to learn the mapping between an input image and an output image using a set of unpaired training images. In this paper, we propose an extension of the unsupervised image-to-image translation problem to multiple input setting. Given a set of paired images from multiple modalities, a transformation is learned to translate the input into a specified domain. For this purpose, we introduce a Generative Adversarial Network (GAN) based framework along with a multi-modal generator structure and a new loss term, \textit{latent consistency loss}. Through various experiments we show that leveraging multiple inputs generally improves the visual quality of the translated images. Moreover, we show that the proposed method outperforms current state-of-the-art unsupervised image-to-image translation methods.   
\end{abstract}

\section{Introduction}

The problem of unsupervised image-to-image translation has made promising strides with the advent of Generative Adversarial Networks (GAN) \cite{goodfellow2014generative} in recent years. Given an input from a particular domain, the goal of image-to-image translation is to transform the input onto a specified second domain. Recent works in image-to-image translation has successfully learned this transformation across various tasks including satellite images to map images, night images to day images, greyscale images to color images etc. \cite{CycleGAN2017}, \cite{NIPS2016_MING}, \cite{NIPS2017_MING} \cite{DualGAN2017}.

\begin{figure}[t]
 	\centering
 	\includegraphics[width=0.8\linewidth]{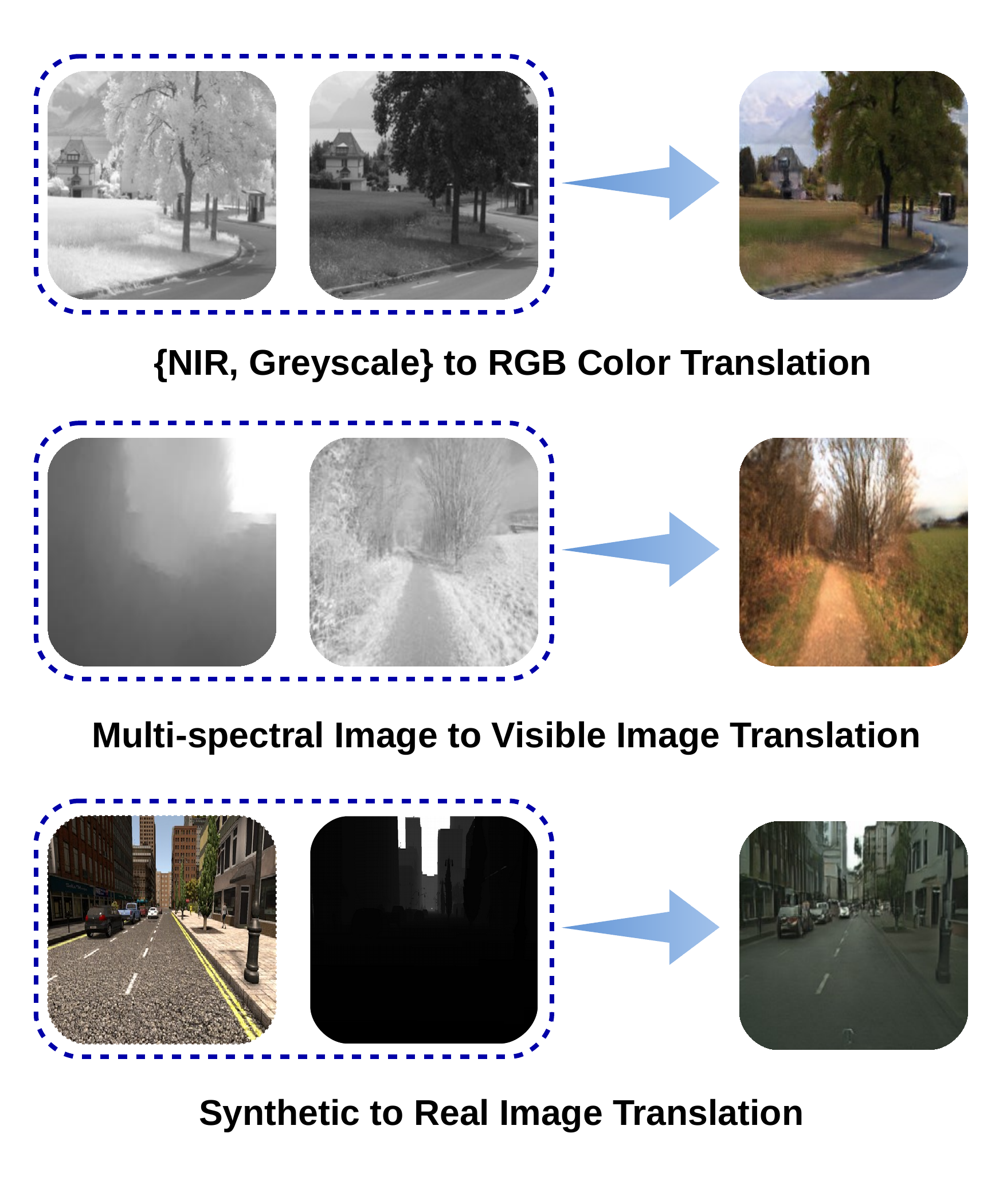} 
 	\vskip -10pt\caption{Traditional unsupervised image-to-image translation maps a single image onto a specified domain. In this work, we address multi-image-to-image translation $(I^{n}2I)$, where a set of input images from different domains are mapped onto a specified domain. This figure illustrates three applications of the proposed method.}\label{fig:mn2m2}
 \end{figure}

In this work, we propose an extension of the original problem from a single input image to multiple input images, called multi-image-to-image translation $(I^{n}2I)$. Given semantically related multiple images across $n$ number of different domains, the goal of $I^{n}2I$ is to produce the corresponding image in a specified domain. For example, the traditional problem of translating a greyscale image onto the RGB domain can be extended into an $I^{n}2I$ problem by providing the near infrared (NIR) image of the same scene as an additional input. Now, the objective would be to use information present in greyscale and NIR domains to produce the corresponding output in the RGB domain as shown in Figure \ref{fig:mn2m2}. In this paper, we study the problem of $I^{n}2I$ in the more generic unsupervised setting and provide initial direction to solve the problem.

Image-to-image translation is a challenging problem. For a given input, there exists multiple possible representations in the specified second domain. Having multiple inputs from different image modalities reduces this ambiguity due to the presence of complimentary information. Therefore, as we show later in experimental results section, leveraging multiple input images leads to an output of higher perceptual quality. 


Multiple input modalities can be incorporated naively by concatenating all available modalities as channels and feeding into an existing image-to-image translation algorithm. However, such an approach leads to unsatisfactory results (see supplementary material for experiments). Therefore, we argue that unsupervised multi-image-to-image translation should be treated as a unique problem.  In this work our main contributions are three-fold:
	 
\noindent 1. We introduce the problem of unsupervised multi-image-to-image translation. We show that by leveraging multiple modalities one can produce a better output in the desired domain as compared to when only a single input modality is used.

\noindent 2. A GAN-based scheme is proposed to combine information of multiple modalities to produce the corresponding output from the desired domain.  We introduce a new \textit{latent consistency loss} term, into the objective function.

\noindent 3. We propose a generalization to the GAN generator network by introducing a multi-modal generator structure.

\section{Related Work}
To the best of our knowledge, $I^n2I$ problem has not been previously addressed in the literature. In this section, we outline previous work related to the proposed method.

\noindent \textbf{Generative Adversarial Networks (GANs).} The fundamental idea behind GAN introduced in \cite{goodfellow2014generative},\cite{NIPS2016_6125} is to use two competing Fully Convolutional Networks (FCN \cite{ShelhamerLD17}), the generator and the discriminator, for generative tasks. Here, the generator FCN learns to generate images from the target domain and the discriminator tries to distinguish generated images from real images of the given domain. During training, two FCNs engage in a min-max game trying to outperform each other.  Learning objective of this problem is collectively called as the \textit{adversarial loss} \cite{CycleGAN2017}. Many applications have since employed GANs for various image generation tasks with success \cite{DCGAN}, \cite{johnson2016perceptual},\cite{SRGAN}, \cite{isola2016image},\cite{Wang_SSGAN2016}, \cite{zhang2016colorful},\cite{han2017stackgan}.

In, \cite{mirza2014conditional} GANs were studied in a conditional setting where a conditioning label is provided as the input to both the generator and the discriminator. Here, we limit our discussion on Conditional GANs (CGAN) to image-to-image generation tasks. The Pix2Pix framework introduced in \cite{isola2016image} uses CGANs for supervised image-to-image translation. In their work, they showed successful transformations across multiple tasks when paired samples across the two domains are available during training. In \cite{cvpr17best}, CGANs are used to generate real eye images from synthetic eye images. In order to learn an effective discriminator, \cite{cvpr17best} proposes to maintain and use a history of generated images. The CoGAN framework introduced in \cite{NIPS2016_MING}, maps images of two different domains onto a common representation to perform domain adaptation using a weight sharing FCN architecture. 

\noindent \textbf{Unpaired image-to-image translation.} Several recent methods have addressed the unsupervised image-to-image translation task when the input is a single image. Here, unlike in the supervised setting, 
paired samples across the two domains do not exist. In \cite{CycleGAN2017}, image-to-image translation problem is tackled by having two generators and discriminators, one for each domain. In addition to the adversarial loss, a cycle consistency constraint is added to ensure that the semantic information is preserved in the translation.  A similar rationale is adopted in DualGAN \cite{DualGAN2017} which has been developed independently of CycleGAN. In \cite{NIPS2017_MING}, the CoGAN framework was extended using GANs and variational autoencoders with the assumption of a common latent space between the domains.

\noindent \textbf{Image fusion.} Although image fusion \cite{Mitchell_Book} operates on multiple input images, we note that our task is very different from image fusion since the former does not involve a domain translation process. In image fusion tasks, multiple input modalities are combined in an informative latent space. This space is usually found by a derived multi-resolution transformation such as wavelets \cite{wavelet}.  In \cite{conf/icml/NgiamKKNLN11} operating on deep networks, a latent space is used to re-generate outputs of multiple modalities. Motivated by this technique, we fuse mid-level deep features from each input domain in the proposed generator FCN.

\section{Proposed Method}
\noindent {\bf{Notation.}} In this paper, we use the following notations. Source domain and target domain are denoted by $S$ and $T$, respectively.  The latent space is denoted by $Z$. In the presence of multiple source domains, the set of source domains $\{S_1,\dots,S_n\}$ are denoted collectively as $\mathcal{S}$. A data sample drawn from an arbitrary domain $X$ is denoted as $x$. The transformation between domains $X$ and $Y$ is denoted by the function $f_{X\rightarrow Y}$. The transformation between the domains $X$ and the latent space $Z$ is denoted by $h_{X \rightarrow Z}$.

\noindent \textbf{Overview.}  In conventional image-to-image translation, the objective is to translate images from an original domain $S$ to a target domain $T$ using a learned transformation $f_{S \rightarrow T}(.)$. In the supervised setting of the problem, a set of image pairs $\{ (s_1,t_1),(s_2,t_2), \dots , (s_p, t_p)\} $ are given, where $s_i \in S $ and $t_i \in T$ are paired images from the two domains. Image-to-image translation task is less challenging in this scenario since the desired output for a given input is known ahead of time.

Similar to the supervised version of the problem, images from both target and source domains are provided in the unsupervised image-to-image translation problem. However, in this case, provided images of the two domains are not paired. In other words, for a given source image $s_i$, the corresponding ground truth image $t_i$ is not provided. In the absence of image pairs from both domains, it is not possible to optimize over a distance between the estimated output and the target. One possible option is to introduce an \textit{adversarial loss} to facilitate reward if the generated image is from the same domain as the target domain.  However, having an adversarial loss alone does not guarantee that the generated image will share semantics with the input. Therefore, to successfully solve this problem, additional constraints need to be imposed. 

\begin{figure}[t]
	\centering
	\includegraphics[width=1\linewidth]{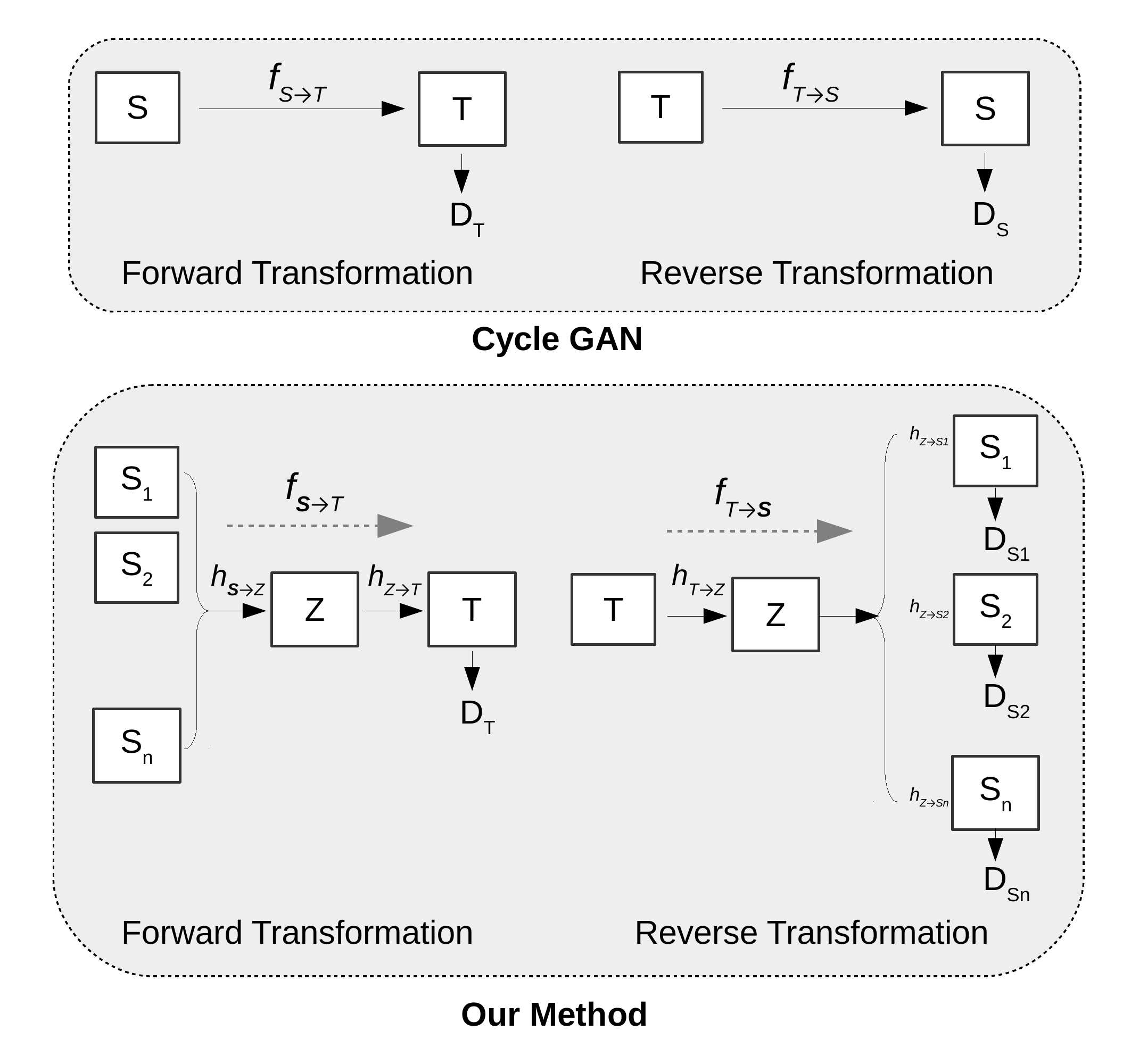} 
	\caption{Network structure used for unsupervised image-to-image translation. Top: CycleGAN, Bottom: Proposed method for $I^n2I$}\label{fig:cycgan}
\end{figure}

In \cite{CycleGAN2017}, such a solution is sought by enforcing the cycle consistency property. Here, an inverse transformation  $f_{T \rightarrow S}(.)$ is learned along with $f_{S \rightarrow T}(.)$. Then, the cycle consistency ensures that the learned transformation yields a good approximation of the input $s_i$ by  comparing $s_i$ with $ f_{ T \rightarrow S}(f_{S \rightarrow T}(s_i))$. We develop our method based on the foundations of CycleGAN proposed in  \cite{CycleGAN2017}. Here, we briefly review the CycleGAN method and we will draw differences between CycleGAN and our method in succeeding sections. CycleGAN as shown in Figure \ref{fig:cycgan} (top), contains a forward transformation from source domain to target domain and a reverse transformation from target to source. Two discriminators $D_S$ and $D_T$ are used to asses whether a given input belongs to source or target, respectively.

\begin{figure}[t]
	\centering
	\includegraphics[width=1\linewidth]{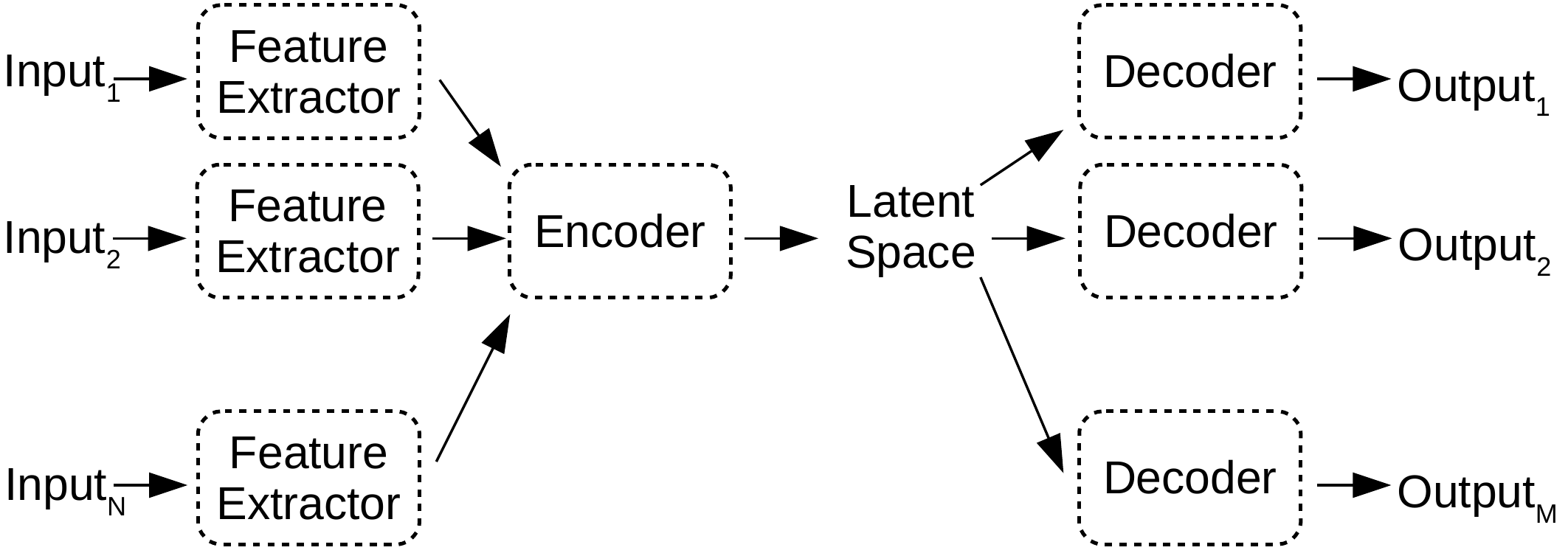} 
	\caption{Multi-modal generator: generalization of the generator for multiple inputs and multiple outputs.}\label{fig:gen}
\end{figure}

\noindent \textbf{Multimodal Generator.}   The $I^n2I$ problem accepts $n$ inputs and translates them into a single output. Therefore, in contrast to CycleGAN, the proposed method deals with multiple inputs in the forward transformation and multiple outputs in the reverse transformation. In order to facilitate this operation, we propose a generalization of the generator structure for multiple inputs and outputs. The generic structure of the proposed generator is shown in Figure \ref{fig:gen}. In general, it is possible for the generator to have $N$ inputs and $M$ outputs. The generator treats each input modality independently and extracts features and fuses them prior to feeding them to the encoder. The encoder maps resultant features to a latent space. Operating on the latent space, $M$ number of independent decoders generate $M$ output images.

For the specific application of $I^n2I$, two generators are used for the forward and reverse transformations. When there are $n$ input images, $M$ is set to be equal to one during the forward transformation where the goal is to generate a single output image ($N=n, M=1$). In the reverse transformation, a single input image is processed to generate $n$ outputs thereby making $N=1$ and $M=n$. Therefore, generator networks used in  $I^n2I$  are asymmetric in structure as shown in Figure \ref{fig:cycgan} (bottom). 

The proposed method treats $n$ inputs independently initially in the forward transformation and then extracted features are fused together. The fused feature is first transformed into a latent space $Z$ as shown in in Figure \ref{fig:cycgan} (bottom) and then transformed into the target domain. In the reverse transformation, the single input is mapped back to the same latent space first. Then, the latent space representation is used to produce $n$ outputs belonging to $n$ source domains. In this formulation, $n+1$ discriminators are used, one for each domain as opposed to CycleGAN. In addition, a latent space consistency loss is added to ensure that the same concept in all domains have a common latent space representation.

\noindent \textbf{Problem Formulation.}
Formally, given $n$ number of input modalities $\mathcal{S} = \{S_1, S_2, \dots , S_n \}$, the objective is to learn a transformation $f_{\mathcal{S} \rightarrow T}(.)$. Here, we note that the input to the forward transformation is a set of images, where the output of the transformation is a single image. Similarly, the backward transformation  $f_{T \rightarrow \mathcal{S}}(.)$ takes a single image input and produces $n$ output images.

In order to approach the solution to this problem, first we view all input images and the desired output image as different representations of the same concept. Motivated by the techniques used in domain adaptation \cite{45545},\cite{GaninUAGLLML16},\cite{sun2015return} we hypothesize the existence of a latent representation that can be derived using the provided representations. With this assumption, we treat our original problem as a series of sub-problems where the requirement is to learn the transformation and the inverse transformation to the latent representation from each domain. If the latent representation is $Z$, we will attempt to learn transformations $h_{I \rightarrow Z}$ and $h_{Z \rightarrow I}$, where $I \in \{\mathcal{S},T\}$  and $ h_{I \rightarrow Z} =  h_{Z \rightarrow I}^{-1}$. With this formulation, the forward transform $f_{\mathcal{S} \rightarrow T}$ becomes $f_{\mathcal{S} \rightarrow T}(.) = h_{Z \rightarrow T}(h_{\mathcal{S} \rightarrow Z}(.))$ and the reverse transformation $f_{T \rightarrow \mathcal{S}}$ becomes $f_{T\rightarrow \mathcal{S}(.) } = h_{Z \rightarrow \mathcal{S}}(h_{T \rightarrow Z}(.))$.

\noindent \textbf{Adversarial Loss.}
In order to learn transformation $f_{\mathcal{S} \rightarrow T}$, we use an adversarial generator-discriminator pair \{$f_{\mathcal{S} \rightarrow T}(.)$, $D_T(.)$\} \cite{goodfellow2014generative}. Denoting the data distributions of domains $\mathcal{S}$ and $T$ as $P_{data}(s)$ and $P_{data}(t)$, respectively, the generator function tries to learn the transformation  $f_{\mathcal{S} \rightarrow T}$. The discriminator is trained to differentiate real images from the target domain $\mathcal{S}$ from generated images $f_{\mathcal{S} \rightarrow T}({s})$. This procedure is captured in the adversarial loss as follows:
\begin{align} 
\label{ganlossf}
\nonumber  L_{GAN,\mathcal{S} \rightarrow T } &= \E_{t\sim p_{data}(t)}[\log D_T (t)] \\&+  \E_{{s} \sim p_{data}({s})}[1-\log \ D_T(f_{\mathcal{S} \rightarrow T}({s}) )]. 
\end{align} 
Similarly, to learn $f_{T \rightarrow S}$ we use a single generator $f_{T \rightarrow S}$. However, since there exists $n$ input domains in total, we require $n$ discriminators $\{D_{s_i}\}$, where $i=1,2,\dots,n$, one for each domain. With this formulation, the total adversarial loss in backward transformation becomes a summation of $n$ adversarial terms as follows:
\begin{dmath} \label{ganlossr} L_{GAN,T \rightarrow \mathcal{S}} =  \sum_{i=1}^{n} \E_{s_i \sim p_{data}(s_i)}[\log D_{S_i}(s_i)]+\sum_{i=1}^{n}
\E_{t\sim p_{data}(t)}[1-\log D_{S_i}  (f_{T \rightarrow S_i}(t))].
 \end{dmath}

\noindent \textbf{Latent Consistency Loss.}
As briefly discussed above, the adversarial loss only ensures that the generated image looks realistic in the target domain. Therefore, adversarial loss alone is inadequate to result in a transformation which preserves semantic information of the input. However, based on the assumption that both input and target domains share a common latent representation, it is possible to enforce a more strict constraint to ensure semantics between the input and the output are preserved. This is done by forcing the latent representation obtained during the forward transformation to be equal to the latent representation obtained during the reverse transformation for the same input. 

More specifically, for a given input ${s}$, a set of latent representations $h_{{s} \rightarrow Z} ({s})$ are recorded. Then, this recorded vector is compared against the latent representation obtained during the reverse transformation $ h_{T \rightarrow Z} (f_{\mathcal{S} \rightarrow T}({s})) $. The latent consistency loss in the forward transformation is defined as,
\begin{dmath} \label{latentlossf}
	L_{latent,\mathcal{S} \rightarrow T} = \E_{s \sim p_{data}({s})} \|h_{\mathcal{S} \rightarrow Z} ({s})-h_{T \rightarrow Z} (f_{\mathcal{S} \rightarrow T}({s})) \|_1.
	\end{dmath}
Similarly, the latent consistency loss in the reverse transformation is defined as,
\begin{dmath} \label{latentr}
	L_{latent,T\rightarrow \mathcal{S} } =  \E_{t \sim p_{data}(t)}
	\|h_{T \rightarrow Z} (t)-h_{\mathcal{S} \rightarrow Z} (f_{T \rightarrow \mathcal{S}}(t)) \|_1.
\end{dmath}

\noindent \textbf{Cycle Consistency Loss.}
 If the input and the transformed image do not share semantic information, it is impossible to regenerate the input using the transformed image. Therefore by forcing the learned transformation to have a valid inverse transform, it is further possible to force the generated image to share semantics with the input. Based on this rationale, in \cite{CycleGAN2017} cycle consistency loss is introduced to ensure that the transformed image shares semantics with the input image. Since this argument is equally valid for the multi-input case, we adopt cycle consistency loss \cite{CycleGAN2017} in our formulation. Proposed backward cycle consistency loss is similar to that of \cite{CycleGAN2017} in definition. We define the reverse cycle consistency loss as:
\begin{dmath} \label{cyclossf} L_{cyc,T \rightarrow \mathcal{S}} = \E_{t\sim p_{data}(t)}[\| f_{\mathcal{S} \rightarrow T}(f_{T \rightarrow \mathcal{S}}(t))-t\|_1].
\end{dmath}
However, in comparison, the forward cycle consistency loss takes into account $n$ inputs and compares the distance among the $n$ reconstructions as opposed to \cite{CycleGAN2017}. The forward cycle consistency loss is defined as,
\begin{dmath} \label{cyclossr} L_{cyc, {s} \rightarrow T} = \E_{{s}\sim p_{data}({s})}[\| F_{T \rightarrow {s}}(F_{{s} \rightarrow T}({s}))-{s}\|_1].  
\end{dmath}

\noindent \textbf{Cumulative Loss.}
The final objective function is the addition of all three losses introduced in this section. The cumulative loss $L_{total}$ is defined as follows:
\begin{dmath} \label{eq:tloss}L_{total} = 
L_{GAN,\mathcal{S} \rightarrow T } +  L_{GAN,T \rightarrow \mathcal{S}} + \lambda_1 ( L_{cyc,T \rightarrow \mathcal{S}} +  L_{cyc, \mathcal{s} \rightarrow T} ) + \lambda_2 (	L_{latent,\mathcal{S} \rightarrow T}+	L_{latent,T\rightarrow \mathcal{S} }), \end{dmath}
where, $\lambda_1$ and $\lambda_2$ are constants. 


\noindent \textbf{Limiting Case.}
It is interesting to investigate the behavior of the proposed network in the limiting case when $n=1$. In this case, both the number of input and output modalities of the network becomes one; i.e. $N=1$ and $M=1$. Therefore $\mathcal{S}$ becomes $S$ in equations \eqref{ganlossf}, \eqref{ganlossr}, \eqref{cyclossf} and \eqref{cyclossr}. In addition, with $n=1$, summation in \eqref{ganlossr} reduces to a single statement. If we disregard the latent consistency loss by forcing $\lambda_2 = 0$, the total objective reduces to, 
\begin{dmath*} 
 L_{total} = \E_{t\sim p_{data}(T)}[\log D_T (t)] +  \E_{s \sim p_{data}(S)}[\log D_{S}(s)]+ \E_{s \sim p_{data}(s)}[1-\log \ D_T(f_{s \rightarrow T}(s) )]+  \E_{t\sim p_{data}(T)}[1-\log D_{S} (t)] +  \E_{t\sim p_{data}(t)}[\| f_{s \rightarrow T}(f_{T \rightarrow s}(t))-t\|_1] \\+\E_{s\sim p_{data}(s)}[\| F_{T \rightarrow s}(F_{s \rightarrow T}(s))-s\|_1]. 
 \end{dmath*} 
This reduced objective is identical to the total objective in CycleGAN. Therefore, in the limiting case when $n=1$, the proposed method reduces to the cycleGAN formulation when the latent consistency loss is disregarded.

\begin{figure}[t]
	\centering
	\includegraphics[width=1\linewidth]{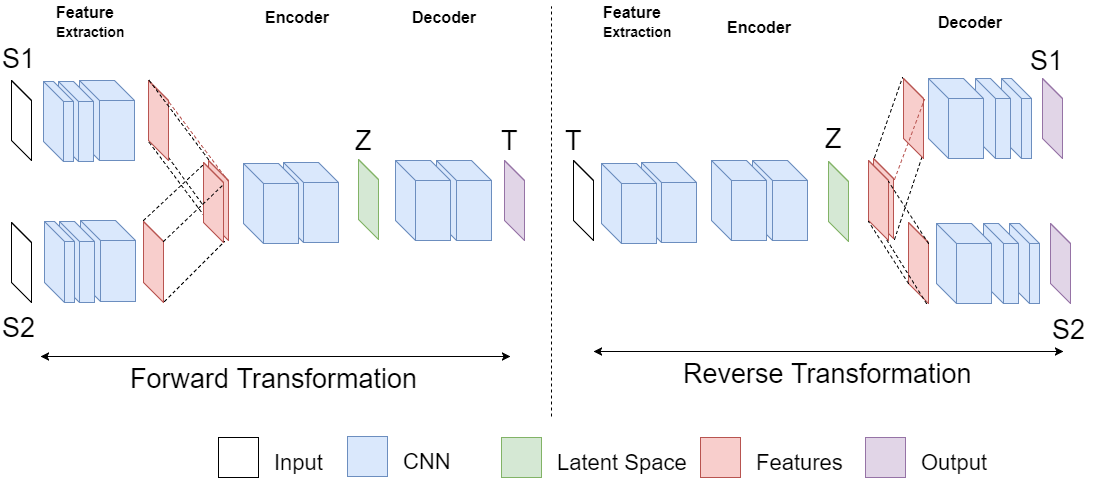}
	\caption{Generator architecture of for $I^n2I$ when two input modalities are used.}\label{fig:mod2}
\end{figure}

\noindent \textbf{Network Architecture.}
In this section, we describe the network architecture of the proposed Generator by considering the case where two input modalities are used; i,e when $n=2$. The resulting two generators in this case is illustrated in Figure~\ref{fig:mod2}. It should be noted that the Convolutional Neural Network (CNN) architectures used in both forward and reverse transformations here are in coherence with the generic structure shown in Figure~\ref*{fig:gen}. In principle, the generator can be based on any backbone architecture. In our work, we used ResNet \cite{HeZRS16} with nine resnet blocks as the backbone. In our proposed network, a CNN is used for each module in Figure~\ref{fig:gen}. These CNNs are typically convolutions/transposed convolutions followed by nonlinearities, batch-normalization layers and possibly with skip connections.  

Two input images (from the two input domains) are present as the input of the forward transformation. These images are subjected to two parallel CNNs to extract features from each modality. Then, the extracted features are fused to generate an intermediate feature representation. In our work, feature fusion is performed by concatenating feature maps of feature extraction stage and using a convolution operation to reduce the dimension. This feature is then subjected to a set of convolution operations to arrive at the latent space. Finally, the latent space representation is subjected to a series of CNNs with transposed convolution operations to generate a single output image (from the target domain). 

During the backward transformation, a single input is present. A CNN with convolution operations is used to transform the input into the latent space. It should be noted that since there is only a single input, there is no notion of fusion in this case. Two parallel CNNs consisting of transposed convolutions branch out from the latent space to produce two outputs corresponding to domains $S1$ and $S2$. 

This architecture can be extended $n$ modalities. In this case, the core structure will be similar to that of Figure~\ref{fig:mod2} except that there will be $n$ parallel branches instead of two at either ends of the network.  For the discriminator networks we use PatchGANs proposed in \cite{isola2016image}.   Please refer to the supplementary material for exact details of the architecture.          

\section{Experimental Results}
We test the proposed method on three publicly available multi-modal image datasets across three tasks against state-of-the-art unsupervised image-to-image translation methods. The training was carried out adhering to principles of unsupervised learning. Even when ground truth images of the desired translation were available, they were not used during training. When available, ground truth images were used during testing to quantify the structural distortion introduced by each method through calculating PSNR and SSIM \cite{Wang:2004:IQA:2319031.2320551} metrics. It should be noted that in this case, two disjoint image sets were used for training and testing.

As the benchmark for performance comparison, we use CycleGAN \cite{CycleGAN2017} and UNIT \cite{NIPS2017_MING} frameworks. Since both of these methods are specifically designed for single inputs, we used all available image modalities, one at a time to produce the corresponding outputs. In the implementation of the proposed method, $\lambda_1$ and $\lambda_2$  in \eqref{eq:tloss} are set equal to 10 and 1, respectively. Learning is performed using the Adam optimizer\cite{adam} with a batch size of 1. Initial learning rates of generators and discriminators were set equal to 0.0002 and 0.0001, respectively. Training was conducted for 200 epochs, where learning rate was linearly decayed in the last 100 epochs.

\noindent \textbf{Image Colorization (EPFL NIR-VIS Dataset.)}
The EPFL NIR-VIS dataset \cite{BS11} includes  477 images in 9 categories captured in the RGB and the Near-infrared (NIR) image modalities across diverse scenes. Scenes included in this dataset are categorized as country, field, forest, indoor, mountain, old building, street, urban and water. We use this dataset to simulate the image colorization task. We generated greyscale images from the RGB visible images and use greyscale and NIR images as the input modalities with the aim of producing the corresponding RGB image. We randomly selected 50 images to be the test images and used the remaining images for training.

\begin{figure*}[htp!]
	\centering
	\includegraphics[width=.8\linewidth]{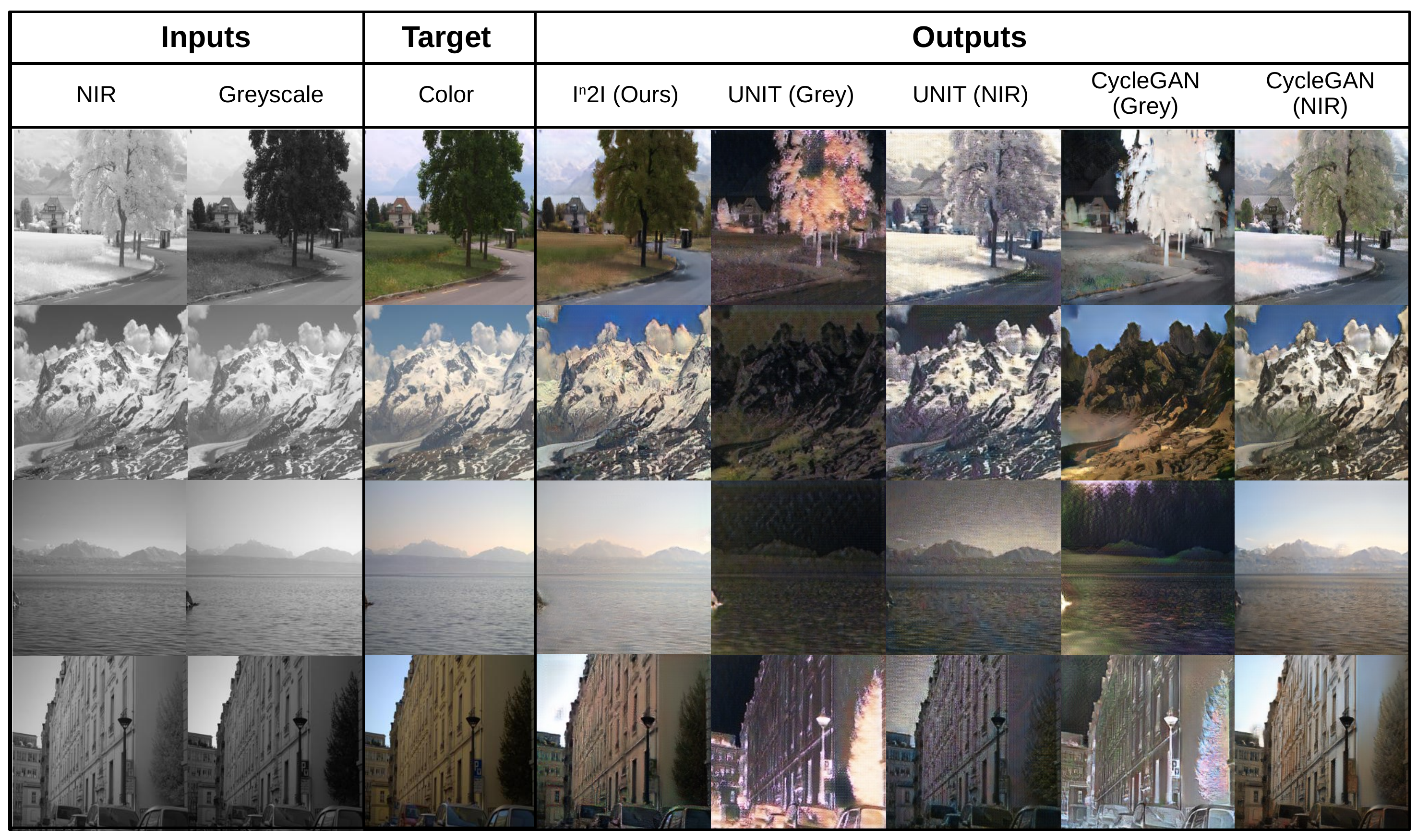} 
	\vskip -5pt\caption{Qualitative results for the image colorization task. }\label{fig:col}
\end{figure*}

First we trained CycleGAN \cite{CycleGAN2017} and UNIT \cite{NIPS2017_MING} models for each input modality independently. Then, the proposed method was used to train a model based on both input modalities. Obtained results for these cases are shown in Figure~\ref{fig:col}. Obtained PSNR and SSIM values for each method on the test data are tabulated in Table~\ref{table:tbl}. By inspection, CycleGAN operating on greyscale images were able to identify segments in the image but failed to assign correct colors. For example, in the first row, the tree is correctly segmented but with a wrong color. In comparison, CycleGAN with NIR images have resulted in a much better colorization. Since the amount of energy a color reflects depends on the wavelength of the color, a NIR signal contains some information about the color of the object. This could be the reason why NIR images have performed better colorization compared to greyscale. The same trend can be observed in the outputs of the UNIT method.

On the other hand, the proposed method has produced a colorization very similar to the ground truth. As an example we wish to draw the attention of the reader to the color of the tree and the field in the first row, colors of the building and the tree in the last row. It has also recorded a superior PSNR and SSIM values compared with the other baselines as shown in Table~\ref{table:tbl}. It should be noted that PSNR and SSIM values only reflect how well the structure of objects in images have been preserved. It is not meant to be an indication of how well colorization task has been carried out.

\begin{table}[htp!]
	\centering
	\small
		\caption{Mean PSNR and SSIM values for image colorization and hyperspectral-to-real image translation tasks. Variance is indicated within the brackets. }
	\label{table:tbl}
	\resizebox{\linewidth}{!}{
	\begin{tabular}{|p{1.5cm} | p{1cm} | p{1cm}|p{1.5cm} | p{1cm} | p{1cm}|}
		\hline
			 \multicolumn{3}{|c|}{EPFL NIR-VIS} & \multicolumn{3}{|c|}{Freiburg Forest Dataset} \\ \hline
		
		 Method & PSNR & SSIM & Method & PSNR & SSIM \\\hline\hline
		 
		 Ours (NIR+Grey)& \textbf{23.113  (9.147) } & \textbf{0.739  (0.008)} & Ours (NIR+Depth) &
		 \textbf{19.444    (7.059)} &    \textbf{0.584    (0.012)}  \\ \hline
		 UNIT (Grey) & 8.324    (2.219)  &  0.041    (0.018) & UNIT (Depth) & 9.681 (1.490) & 0.414 (0.007) \\ \hline
		 UNIT (NIR) & 15.331    (9.088)  & 0.544    (0.012) & UNIT (NIR) & 9.494 (0.868)& 0.382 (0.004) \\ \hline
	
	 CycleGAN (Grey) &8.438 (2.939)   &  0.056 (0.018)  & CycleGAN (Depth) & 16.5945    (4.308) &   0.525    (0.010) \\ \hline
 CycleGAN (NIR) &17.381 (9.345)  &  0.657 (0.018) &  CycleGAN (NIR) &18.574   (3.252)   & 0.552    (0.014) \\ \hline 
 - & -  &  - &  Ours (All Inputs) &\textbf{21.65 (2.302)}   & \textbf{0.600    (0.0105)} \\
	\hline
		
	\end{tabular} 
}
\end{table}

\noindent \textbf{Synthetic-to-Real Image Translation (Synthia and CityScapes Datasets).}
In this subsection, we experiment on generating real images using synthetic images. For this purpose, we use two datasets, Synthia \cite{RosCVPR16} and CityScapes \cite{Cordts2016Cityscapes}, respectively as the source and the target domains.  The Cityscapes dataset contains images taken across fifty urban cities during daytime. We use 1525 images from the validation set of the dataset to represent the target domain in the synthetic-to-real translation task.  The Synthia dataset contains graphical simulations of an urban city. The scenes included in the dataset contain different weather conditions, lighting conditions and seasons. For our work, we only use the summer day light subset of the dataset which includes 901 images for training. The Synthia dataset provides RGB image intensities as well as the depth information of the scene.  Hence, we use these as the two input modalities.

\begin{figure*}[htp!]
	\centering
	\includegraphics[width=.8\linewidth]{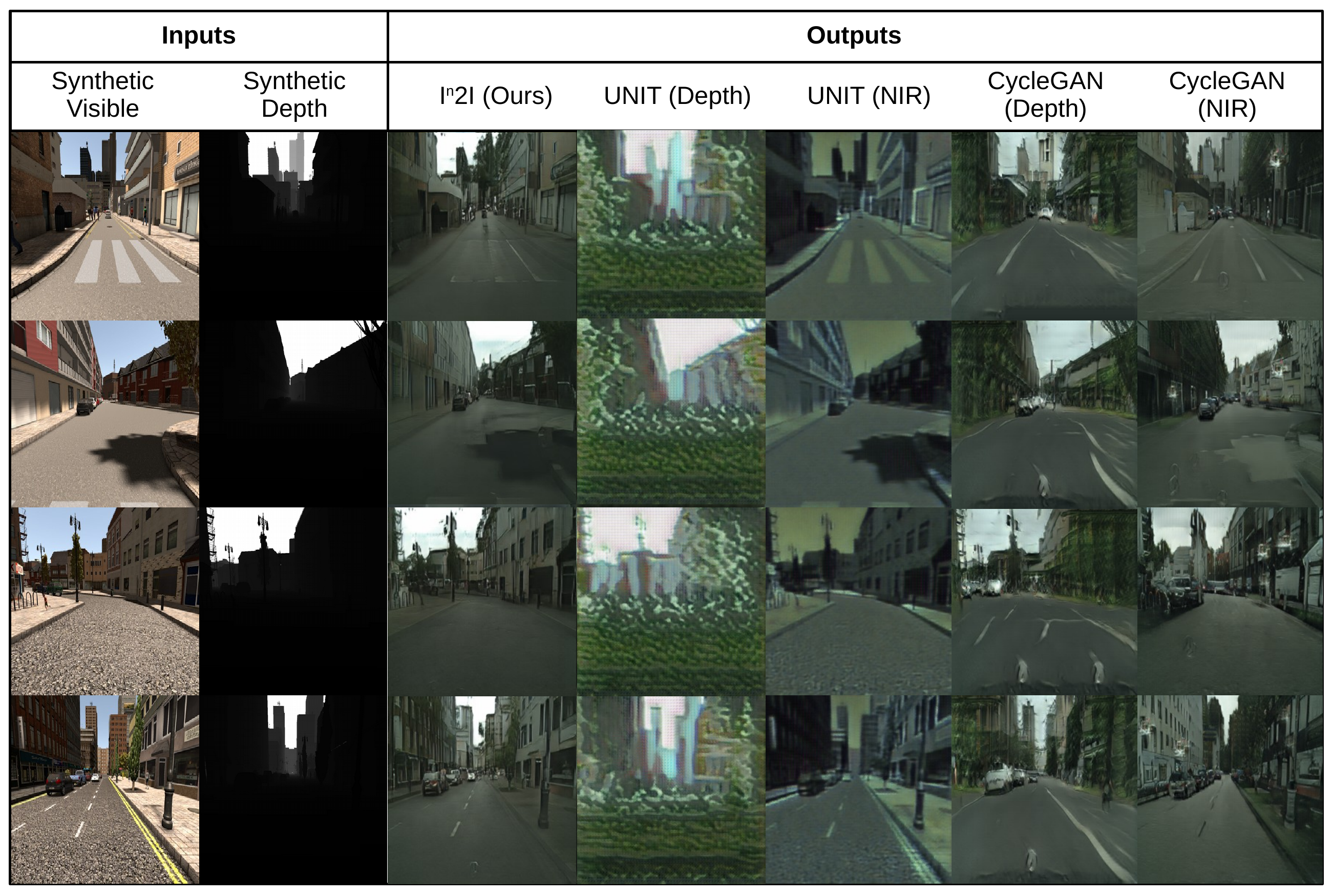} 
	\vskip-5pt\caption{Qualitative results for the synthetic-to-real translation task. }\label{fig:syn2real}
\end{figure*}

Results of this experiment are shown in Figure~\ref{fig:syn2real}. In this particular task, UNIT method has only changed the generic color scheme of the scene with incorrect association; for example note that skies look brown instead of blue in resulting images. In addition, objects in the scene continues to possess the characteristics of synthetic images. In contrast, CycleGAN has attempted to convert appearance of synthetic images to real. However, in the process it has distorted the structure of objects. When only depth information is used, the cycleGAN method is unable to preserve the structure of objects in the scene. For example,  lines along the roads have ended up being warped in the learned representation in Figure~\ref{fig:syn2real}. The CycleGAN model based on the RGB images preserves the overall structure to an extent. However, vital details are either missing or misleading. For  example, pavements are missing from images shown in rows 2 and 3 in Figure~\ref{fig:syn2real}. The absence of a shadow on the road in row 2, addition of clutter in the left pavement in row 3 and disappearance of the telephone pole in row 4 are some of the notable incoherences. Comparatively, fusion of both RGB and depth information has resulted in a more realistic translation. It should be noted that synthetic-to-real translation is a challenging problem in practice and when certain concepts were missing in either of source or target domains, the model found it difficult to learn such concepts. For example, training images from Cityscape did not have zebra crossings in any of the images. Therefore, the concept of zebra crossings is not learned well by the model as shown in row 1.

\begin{figure*}[htp!]
	\centering
	\includegraphics[width=0.9\linewidth]{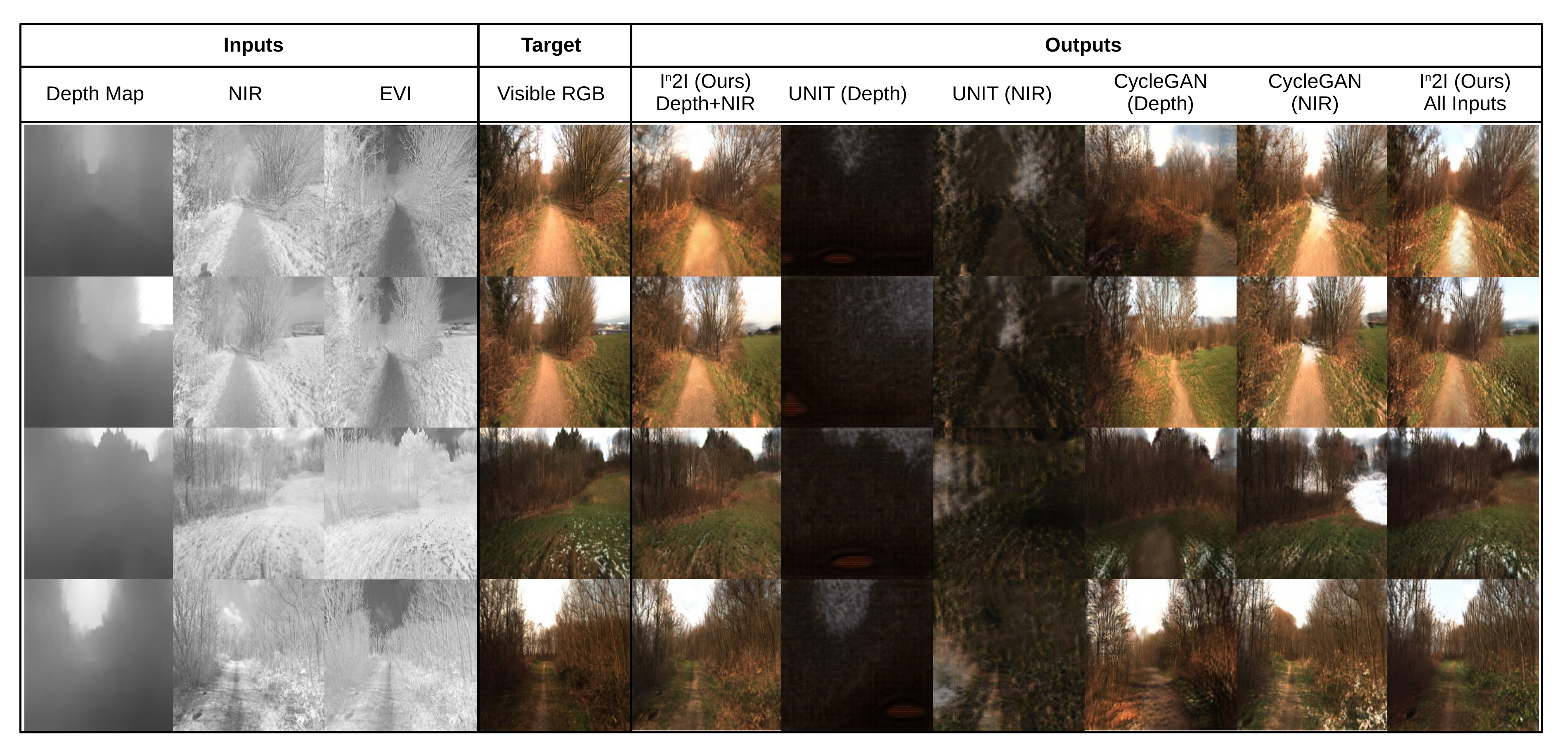} 
	\vskip-5pt\caption{Qualitative results for the hyperspectral-to-visible translation. }\label{fig:for}
\end{figure*}

\noindent \textbf{Hyperspectral-to-Visible Image Translation.}
The Freiburg Forest Dataset \cite{Valada_2016_ISER} contains 230 training images and 136 test images of a forest scenery with images of different wavelengths. The types of images contained in the dataset include RGB, depth, NIR, EVI (Enhanced Vegetation Index) and combinations of aforementioned image types. In our experiments, we use this dataset to perform hyperspectral-to-real image translation where the depth and the NIR image modalities are used to recover the RGB images of the visual spectrum. 

We note that this task is relatively easier compared to the earlier two tasks due to the low diversity in scenes. All methods, except for UNIT, were able to generate realistic RGB images from the provided input modalities, as shown in Figure~\ref{fig:for}. The reason why UNIT fails in this task could be due to the low number of training image samples \cite{NIPS2017_MING}. In CycleGAN, visual domain image reconstruction solely based on the depth information has resulted in sub-standard images. For example, the road is missing in row 1 and the road takes a wrong shape in row 2 for the depth image-based CycleGAN output. When the NIR images are used as the input, the resulting CycleGAN output is more closer to the ground truth. But, in this case, there exists multiple missing regions, where pixels are painted in white, as shown in rows 1-3.  

In terms of reconstructing finer details, the proposed method that utilizes both NIR and depth information have outperformed the other baselines. In comparison to the earlier task, all methods have less structural distortion as evident from Table~\ref{table:tbl}. However, even in this case, the proposed method has performed marginally better than the other baseline methods in terms of SSIM and PSNR performance. 

Since the dataset has three modalities available, we ran an extra experiment by inputting all three modalities (Depth, NIR and EVI) into the proposed method. Results produced in this case were more closer to the ground truth in general. For example, water residues on the ground are produced in row 3 for this case, which was absent in all prior cases. The use of three modalities have improved the PSNR performance by more than $2.0$ according as shown in Table~\ref{table:tbl}. Please refer to supplementary material for more results and analysis.

\begin{figure}[htp!]
	\centering
	\includegraphics[width=0.9\linewidth]{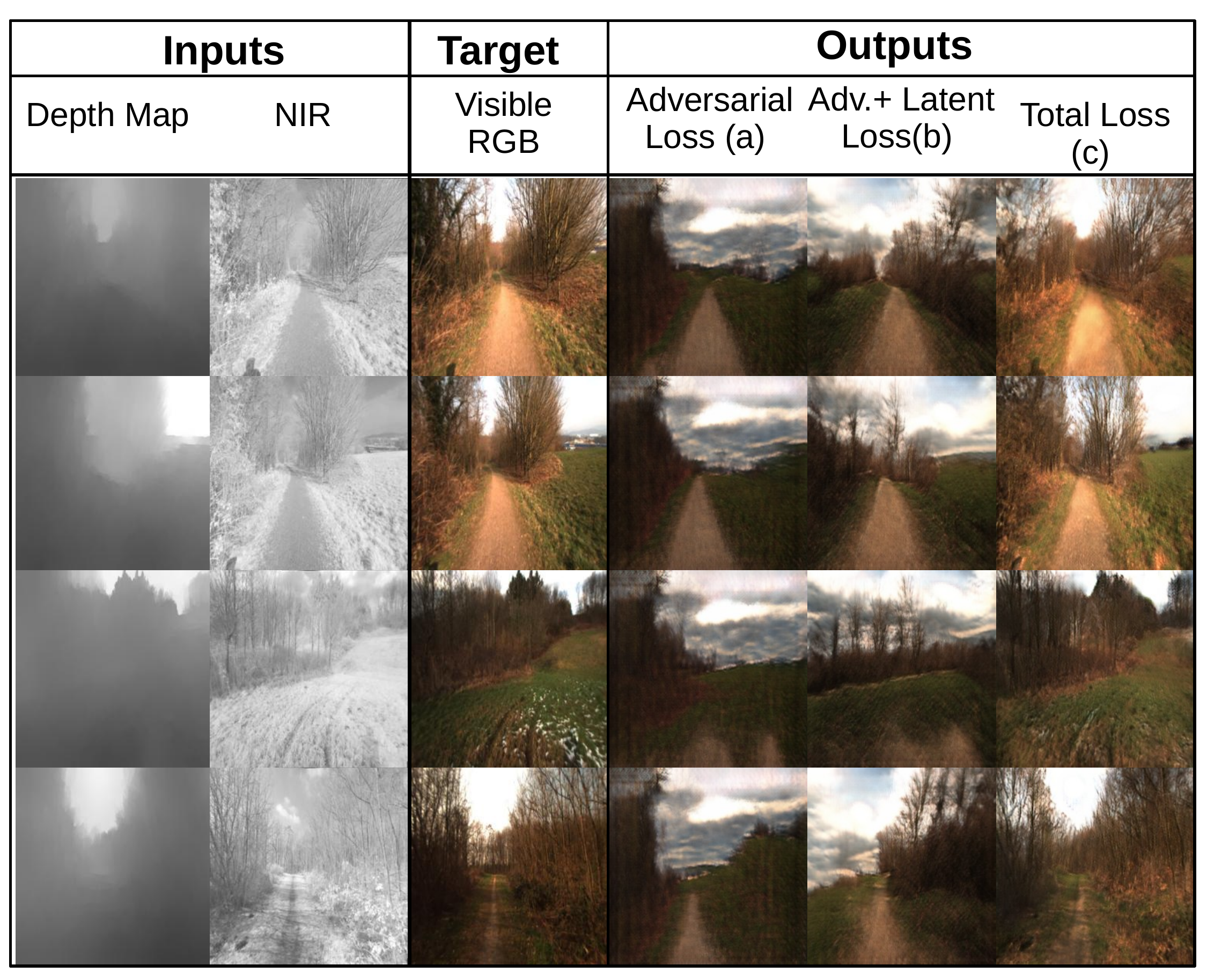} 
	\vskip -5pt\caption{Results of the ablation study carried out on the hyperspectral-to-visible translation task. }\label{fig:ablation}
\end{figure}

\noindent \textbf{Ablation Study.}
The loss function proposed in \eqref{eq:tloss} has three main components: the adversarial loss, latent consistency loss and cycle consistency loss. In this subsection, we carry out an ablation study on the Freiburg Forest Dataset to investigate the impact of each individual loss term. In this study, we considered three alternative loss functions: (a) only adversarial loss, (b) addition of adversarial loss and the latent consistency loss and the (c) total loss. Obtained image constructions for each case for a set of sample images are shown in Figure~\ref{fig:ablation}. Images generated in case (a) are plausible forest images; but they are very different from the ground truth. This is because the adversarial loss doesn't take semantic information into account. Comparatively, images in case (b) shares more semantics with ground truth. For example, in row 1, some trees are generated at the right side of the road when compared with case (a). Further, image artifacts present at the left side in case (b) have  disappeared in case (c). Addition of the cycle consistency loss in case (c), increases the coherence between the output and the ground truth even more compared with the earlier cases. 

\noindent \textbf{Impact of Multiple Inputs.}
Three experiments performed in this section are of different levels of difficulty. The hyperspectral-to-visible image translation task is the easiest task due to the low variance in the scenes in the dataset. In such scenarios, even a single modality is able to produce a reasonable translation. However, we note that introducing an additional modality has improved the performance. In comparison, the colorization task is more challenging due to the availability of diverse scenes. As a result, a single modality was not able to perform colorization satisfactorily. In this case, multi-image-to-image translation was able to induce a high improvement in terms of visual quality by using two informative input modalities. The final case, synthetic-to-real image translation, is very challenging. We note that the depth modality in this case is not very informative since it leads to image constructions of sub-standard quality. In comparison, the RGB synthetic image modality resulted in better translations. Using both modalities has improved the visual quality of the output. But this improvement was marginal as compared to the case of the colorization task.

In summary, multiple modalities generally improve the visual quality of the output image; specially when the translation is more challenging. However, the amount of improvement introduced was dependent on the informativeness of the second modality. In fact, introducing a noisy modality for the sake of having multiple inputs would not contribute towards an improvement. 


\section{Conclusion}
In this work, we introduced multi-image-to-image translation problem. We proposed a multi-modal generator structure and a GAN based framework as the initial direction to solve the problem. We tested the proposed method across three tasks against state-of-the-art unsupervised image-to-image translation methods. We showed that using multiple image modalities improves the visual quality of the output compared with results generated by the state-of-the-art methods. We analyzed the behavior of the proposed method in the limiting case and provided discussion as to when the use of multiple image modalities is most suited.  

\section{Supplimetary Materials}

\subsection{Detailed Network Architecture}
In this section, we provide details about the architecture used in the generator network for the cases when two and three input modalities are used. In both cases there are two generators; one for the forward transformation and the other for the reverse transformation.  These generators follow the structure shown in Figure 3 in the main paper. 

When two input modalities are available, the used generator structure is shown in Figure 4 in the paper in modular form. Exact details of each of these modules are tabulated in Table~\ref{table:tbl2f}. Details of the reverse transformation network shown in Figure 4 in the paper can be found in Table~\ref{table:tbl2r}. Here, \textit{Conv} refers to a collection of convolution, instance normalization and relu layers.   \textit{Deconv} refers to a collection of transposed convolution, instance normalization and relu layers. \textit{Res} refers to a Resnet block.

\begin{table}[htp!]
	\centering
	\small
	\caption{Network details corresponding to the forward transformation generator for $n=2$.}
	\label{table:tbl2f}
	\resizebox{\linewidth}{!}{
		\begin{tabular}{| c | p{1.5cm} | p{1.3cm}|p{1.3cm} | p{2.5cm}| p{1cm}|}
			\hline
			& Layer & Input & output & Kernel & (stride, pad)  \\\hline\hline
			
			\multirow{7}{*}{Feature Extraction $S_1$}  & Conv  $S_11$ & Image 1 & Conv $S_12$  & $1 \times 7 \times 7 \times 64$  & (1,0) \\

			& Conv $S_12$ &  Conv  $S_11$  & Conv $S_13$ & $1 \times 3 \times 3 \times 128 $  & (2,1) \\

			& Conv $S_13$ &  Conv  $S_12$  & Res $S_11$ & $1 \times 3 \times 3 \times 256 $  & (2,1) \\
			
			& Res $S_11$ &  Conv $S_13$ & Res $S_12$ & $1 \times 3 \times 3 \times 512 $  & (1,0) \\
			
			& Res $S_12$ & Res $S_11$  & Res $S_13$ & $1 \times 3 \times 3 \times 1024 $  &  (1,0)\\
			
			& Res $S_13$& Res $S_12$  & Res $S_14$ & $1 \times 3 \times 3 \times 1024 $  & (1,0) \\

			& Res $S_14$& Res $S_13$  & Conv $F$  & $1 \times 3 \times 3 \times 1024 $  &  (1,0)\\ \hline

			\multirow{7}{*}{Feature Extraction $S_2$}  & Conv  $S_21$ & Image 2 & Conv $S_22$  & $1 \times 7 \times 7 \times 64$  & (1,0) \\

			& Conv $S_22$ &  Conv  $S_21$  & Conv $S_23$ & $1 \times 3 \times 3 \times 128 $  & (2,1) \\
			
			& Conv $S_23$ &  Conv  $S_22$  & Res $S_21$ & $1 \times 3 \times 3 \times 256 $  & (2,1) \\
			
			& Res $S_21$ &  Conv $S_23$ & Res $S_22$ & $1 \times 3 \times 3 \times 512 $  &  (1,0)\\
			
			& Res $S_22$ & Res $S_21$  & Res $S_23$ & $1 \times 3 \times 3 \times 1024 $  & (1,0) \\
			
			& Res $S_23$& Res $S_22$  & Res $S_24$ & $1 \times 3 \times 3 \times 1024 $  &  (1,0)\\

			& Res $S_24$& Res $S_23$  & Conv $F$ & $1 \times 3 \times 3 \times 1024 $  &  (1,0)\\ \hline

			\multirow{2}{*}{ Fusion}	& Conv $F$ & Res $S_13$, Res $S_23$  &  Res $3$ & $1 \times 3 \times 3 \times 1024 $  & (1,0)  \\ \hline
			
			\multirow{4}{*}{Encoder}	& Res $3$& Conv $F$  & res $4$ & $1 \times 3 \times 3 \times 1024 $  & (1,0)  \\ 
			
			& Res $4$& Res $3$  & res $5$ & $1 \times 3 \times 3 \times 1024 $  & (1,0)  \\ 
			
			& Res $5$& Res $4$  & res $6$ & $1 \times 3 \times 3 \times 1024 $  & (1,0)  \\ 
			
			& Res $6$& Res $5$  & Latent & $1 \times 3 \times 3 \times 1024 $  & (1,0)  \\ \hline
			
			\multirow{1}{*}{Z} & Latent & Res $6$  & Res $7$ & Identity  &  \\  \hline
			
			\multirow{6}{*}{Decoder} & Res $7$& Latent  & res $8$ & $1 \times 3 \times 3 \times 1024 $  &  (1,0)  \\ 
			
			& Res $8$& Res $7$  & res $9$ & $1 \times 3 \times 3 \times 512 $  &  (1,0)  \\ 
			
			& Res $9$& Res $8$  & Deconv $1$ & $1 \times 3 \times 3 \times 256 $  &  (1,0) \\ 
			
			& Deconv $1$& Res $9$  & Deconv $2$ & $1 \times 3 \times 3 \times 128 $  & (2,1) \\

			& Deconv $2$& Deconv $1$  & Deconv $3$ & $1 \times 3 \times 3 \times 64 $  & (2,1) \\ 
			
			& Deconv $3$& Deconv $2$  & Output & $1 \times 7 \times 7 \times 3 $  & (1,0) \\ \hline

		\end{tabular} 
	}
\end{table}

\begin{table}[htp!]
	\centering
	\small
	\caption{Network details corresponding to the reverse transformation generator for $n=2$.}
	\label{table:tbl2r}
	\resizebox{\linewidth}{!}{
		\begin{tabular}{|c | p{1.5cm} | p{1.3cm}|p{1.3cm} | p{2.5cm}| p{1cm}|}
			\hline
			& Layer & Input & output & Kernel & (stride, pad)  \\\hline\hline

			\multirow{3}{*}{Feature Extraction }  & Conv  $1$ & Image 1 & Conv $2$  & $1 \times 7 \times 7 \times 64$  & (1,0) \\

			& Conv $2$ &  Conv  $1$  & Conv $3$ & $1 \times 3 \times 3 \times 128 $  & (2,1) \\
			
			& Conv $3$ &  Conv  $2$  & Res $1$ & $1 \times 3 \times 3 \times 256 $  & (2,1) \\
			
			\hline
			
			& Res $1$ &  Conv $2$ & Res $2$ & $1 \times 3 \times 3 \times 512 $  & (1,0) \\ 
			
			\multirow{4}{*}{ Encoder}	& Res $2$ & Res $1$  & Res $3$ & $1 \times 3 \times 3 \times 1024 $  & (1,0) \\
			
			& Res $3$& Res $2$  & Res $4$ & $1 \times 3 \times 3 \times 1024 $  &  (1,0) \\

			& Res $4$& Res $3$  & Latent & $1 \times 3 \times 3 \times 1024 $  & (1,0) \\ \hline
			Z & Latent & Res $4$  & Res $S_15$, Res $S_25$ & Identity  &  \\  \hline
			
			\multirow{12}{*}{Decoder $S_1$}	& Res $S_15$ & Latent  & res $S_16$ & $1 \times 3 \times 3 \times 1024 $  & (1,0) \\ 
			
			& Res $S_16$& Res $S_15$  & Res$S_17$& $1 \times 3 \times 3 \times 1024 $  &  (1,0)\\  
			
			& Res $S_17$& Res $S_16$  & res $S_18$ & $1 \times 3 \times 3 \times 1024 $  & (1,0) \\ 
			
			& Res $S_18$& Res $S_17$  & Res $S_19$ & $1 \times 3 \times 3 \times 512 $  & (1,0) \\
			& Res $S_19$& Res $S_18$  & Deconv $S_11$ & $1 \times 3 \times 3 \times 256 $  &  (1,0) \\

			& Deconv $S_11$& Res $S_19$  & Deconv $S_12$ & $1 \times 3 \times 3 \times 128 $  & (2,1) \\
			
			& Deconv $S_12$& Deconv $S_11$  & Deconv $S_13$ & $1 \times 3 \times 3 \times 64 $  & (2,1) \\
			
			& Deconv $S_13$& Deconv $S_12$  & $Output S_1$ & $1 \times 7 \times 7 \times 3 $  & (1,0) \\ \hline
			
			\multirow{12}{*}{Decoder $S_2$}	& Res $S_25$ &Latent  & res $S_26$ & $1 \times 3 \times 3 \times 1024 $  &  (1,0) \\ 
			
			& Res $S_26$& Res $S_25$  & Res$S_27$& $1 \times 3 \times 3 \times 1024 $  & (1,0) \\  
			
			& Res $S_27$& Res $S_26$  & res $S_28$ & $1 \times 3 \times 3 \times 1024 $  & (1,0) \\ 
			
			& Res $S_28$& Res $S_27$  & Res $S_29$ & $1 \times 3 \times 3 \times 512 $  & (1,0) \\ 
			& Res $S_29$& Res $S_28$  & Deconv $S_21$ & $1 \times 3 \times 3 \times 256 $  & (2,1) \\

			& Deconv $S_21$& Res $S_29$  & Deconv $S_22$ & $1 \times 3 \times 3 \times 128 $  & (2,1) \\

			& Deconv $2_22$& Deconv $S_21$  & Deconv $S_23$ & $1 \times 3 \times 3 \times 64 $  & (2,1) \\

			& Deconv $S_23$& Deconv $S_22$  & $Output S_1$ & $1 \times 7 \times 7 \times 3 $  & (1,0) \\ \hline

		\end{tabular} 
	}
\end{table}

We carried out a single experiment using three modalities using the hyperspectral-to-visible image translation task. We outline the network architecture in both the forward and the reverse transformations in Tables \ref{table:tbl3f} and \ref{table:tbl3r}, respectively.

\begin{table}[htp!]
	\centering
	\small
	\caption{Forward transformation for $n=3$ }
	\label{table:tbl3f}
	\resizebox{\linewidth}{!}{
		\begin{tabular}{|c | p{1.5cm} | p{1.3cm}|p{1.3cm} | p{2.5cm}| p{1cm}|}
			\hline
			& Layer & Input & output & Kernel & (stride, pad)  \\\hline\hline
			
			\multirow{7}{*}{Feature Extraction $S_1$} & Conv  $S_11$ & Image 1 & Conv $S_12$  & $1 \times 7 \times 7 \times 64$  & (1,0) \\
			
			& Conv $S_12$ &  Conv  $S_11$  & Conv $S_13$ & $1 \times 3 \times 3 \times 128 $  & (2,1) \\	
			& Conv $S_13$ &  Conv  $S_12$  & Res $S_11$ & $1 \times 3 \times 3 \times 256 $  & (2,1) \\
			
			& Res $S_11$ &  Conv $S_12$ & Res $S_12$ & $1 \times 3 \times 3 \times 512 $  & (1,0) \\
			
			& Res $S_12$ & Res $S_11$  & Res $S_13$ & $1 \times 3 \times 3 \times 1024 $  & (1,0) \\
			
			& Res $S_13$& Res $S_12$  & Res $S_14$ & $1 \times 3 \times 3 \times 1024 $  & (1,0) \\

			& Res $S_14$& Res $S_13$  & Conv $F$  & $1 \times 3 \times 3 \times 1024 $  & (1,0) \\ \hline

			\multirow{7}{*}{Feature Extraction $S_2$} & Conv  $S_21$ & Image 2 & Conv $S_22$  & $1 \times 7 \times 7 \times 64$  & (1,0) \\

			& Conv $S_22$ &  Conv  $S_21$  & Conv $S_23$ & $1 \times 3 \times 3 \times 128 $  & (2,1) \\	
			& Conv $S_23$ &  Conv  $S_22$  & Res $S_21$ & $1 \times 3 \times 3 \times 256 $  & (2,1) \\

			& Res $S_21$ &  Conv $S_22$ & Res $S_22$ & $1 \times 3 \times 3 \times 512 $  & (1,0) \\
			
			& Res $S_22$ & Res $S_21$  & Res $S_23$ & $1 \times 3 \times 3 \times 1024 $  & (1,0) \\
			
			& Res $S_23$& Res $S_22$  & Res $S_24$ & $1 \times 3 \times 3 \times 1024 $  & (1,0) \\

			& Res $S_24$& Res $S_23$  & Conv $F$ & $1 \times 3 \times 3 \times 1024 $  & (1,0) \\ \hline
			
			\multirow{7}{*}{Feature Extraction $S_3$} & Conv  $S_31$ & Image 3 & Conv $S_32$  & $1 \times 7 \times 7 \times 64$  & (1,0) \\

			& Conv $S_32$ &  Conv  $S_31$  & Conv $S_33$ & $1 \times 3 \times 3 \times 128 $  & (2,1) \\	
			& Conv $S_33$ &  Conv  $S_32$  & Res $S_31$ & $1 \times 3 \times 3 \times 256 $  & (2,1) \\
			
			& Res $S_31$ &  Conv $S_32$ & Res $S_32$ & $1 \times 3 \times 3 \times 512 $  & (1,0) \\
			
			& Res $S_32$ & Res $S_31$  & Res $S_33$ & $1 \times 3 \times 3 \times 1024 $  & (1,0) \\
			
			& Res $S_33$& Res $S_32$  & Res $S_34$ & $1 \times 3 \times 3 \times 1024 $  & (1,0) \\

			& Res $S_34$& Res $S_33$  & Conv $F$  & $1 \times 3 \times 3 \times 1024 $  & (1,0) \\ \hline

			\multirow{3}{*}{Fusion}	& Conv $F$ & Res $S_13$, Res $S_23$, Res $S_33$  &  Res $3$ & $1 \times 3 \times 3 \times 1024 $  & (1,0)  \\ \hline
			
			\multirow{4}{*}{Encoder}	& Res $3$& Conv $F$  & res $4$ & $1 \times 3 \times 3 \times 1024 $  & (1,0) \\ 
			
			& Res $4$& Res $3$  & res $5$ & $1 \times 3 \times 3 \times 1024 $  & (1,0) \\ 
			
			& Res $5$& Res $4$  & res $6$ & $1 \times 3 \times 3 \times 1024 $  & (1,0) \\ 
			
			& Res $6$& Res $5$  & Latent & $1 \times 3 \times 3 \times 1024 $  & (1,0) \\ \hline
			
			Z & Latent & Res $6$  & Res $7$ & Identity  &\\  \hline
			
			\multirow{6}{*}{Decoder} & Res $7$& Latent  & res $8$ & $1 \times 3 \times 3 \times 1024 $  & (1,0) \\ 
			
			& Res $8$& Res $7$  & res $9$ & $1 \times 3 \times 3 \times 512 $  & (1,0) \\ 
			
			& Res $9$& Res $8$  & Deconv $1$ & $1 \times 3 \times 3 \times 256 $  & (1,0) \\ 
			
			& Deconv $1$& Res $9$  & Deconv $2$ & $1 \times 3 \times 3 \times 128 $  & (2,1) \\
			
			& Deconv $2$& Deconv $1$  & Deconv $3$ & $1 \times 3 \times 3 \times 64 $  & (2,1) \\

			& Deconv $3$& Deconv $2$  & Output & $1 \times 7 \times 7 \times 3 $  & (1,0) \\ \hline
		\end{tabular} 
	}
\end{table}

\begin{table}[htp!]
	\centering
	\small
	\caption{Reverse transformation for $n=3$ }
	\label{table:tbl3r}
	\resizebox{\linewidth}{!}{
		\begin{tabular}{|c | p{1.5cm} | p{1.5cm}|p{1.5cm} | l | p{0.7cm}|}
			\hline
			& Layer & Input & output & Kernel & (stride, pad)  \\\hline\hline

			\multirow{3}{*}{Feature Extraction}  & Conv  $1$ & Image 1 & Conv $2$  & $1 \times 7 \times 7 \times 64$  & (1,0) \\

			& Conv $2$ &  Conv  $1$  & Conv $3$ & $1 \times 3 \times 3 \times 128 $  & (2,1) \\
			& Conv $3$ &  Conv  $2$  & Res $1$ & $1 \times 3 \times 3 \times 256 $  & (2,1) \\
			
			\hline
			
			& Res $1$ &  Conv $3$ & Res $2$ & $1 \times 3 \times 3 \times 512 $  & (1,0) \\ 
			
			\multirow{4}{*}{Encoder}	& Res $2$ & Res $1$  & Res $3$ & $1 \times 3 \times 3 \times 1024 $  & (1,0) \\
			
			& Res $3$& Res $2$  & Res $4$ & $1 \times 3 \times 3 \times 1024 $  & (1,0) \\

			& Res $4$& Res $3$  & Latent & $1 \times 3 \times 3 \times 1024 $  & (1,0) \\ \hline
			\multirow{3}{*}{Z} & Latent & Res $4$  & Res $S_15$, Res $S_25$, Res $S_35$ & Identity  &  \\  \hline
			
			\multirow{12}{*}{Decoder $S_1$}	& Res $S_15$ & Latent  & res $S_16$ & $1 \times 3 \times 3 \times 1024 $  & (1,0) \\ 
			
			& Res $S_16$& Res $S_15$  & Res$S_17$& $1 \times 3 \times 3 \times 1024 $  & (1,0) \\  
			
			& Res $S_17$& Res $S_16$  & res $S_18$ & $1 \times 3 \times 3 \times 1024 $  & (1,0) \\ 
			
			& Res $S_18$& Res $S_17$  & Res $S_19$ & $1 \times 3 \times 3 \times 512 $  & (1,0) \\
			& Res $S_19$& Res $S_18$  & Deconv $S_11$ & $1 \times 3 \times 3 \times 256 $  & (1,0) \\

			& Deconv $S_11$& Res $S_19$  & Deconv $S_12$ & $1 \times 3 \times 3 \times 128 $  & (2,1) \\
			
			& Deconv $S_12$& Deconv $S_11$  & Deconv $S_13$ & $1 \times 3 \times 3 \times 64 $  & (2,1) \\ 		
			
			& Deconv $S_13$& Deconv $S_12$  & Output $S_1$ & $1 \times 7 \times 7 \times 3 $  & (2,1) \\ \hline
			
			\multirow{12}{*}{Decoder $S_2$}	& Res $S_25$ &Latent  & res $S_26$ & $1 \times 3 \times 3 \times 1024 $  & (1,0) \\ 
			
			& Res $S_26$& Res $S_25$  & Res$S_27$& $1 \times 3 \times 3 \times 1024 $  & (1,0) \\  
			
			& Res $S_27$& Res $S_26$  & res $S_28$ & $1 \times 3 \times 3 \times 1024 $  & (1,0) \\ 
			
			& Res $S_28$& Res $S_27$  & Res $S_29$ & $1 \times 3 \times 3 \times 512 $  & (1,0) \\ 
			& Res $S_29$& Res $S_28$  & Deconv $S_21$ & $1 \times 3 \times 3 \times 256 $  & (1,0) \\

			& Deconv $S_21$& Res $S_29$  & Deconv $S_22$ & $1 \times 3 \times 3 \times 128 $  & (2,1) \\
			
			& Deconv $S_22$& Deconv $S_21$  & Deconv $S_23$ & $1 \times 3 \times 3 \times 64 $  & (2,1) \\ 		
			
			& Deconv $S_23$& Deconv $S_22$  & Output $S_2$ & $1 \times 7 \times 7 \times 3 $  & (2,1) \\ \hline

			\multirow{12}{*}{Decoder $S_3$}	& Res $S_35$ &Latent  & res $S_36$ & $1 \times 3 \times 3 \times 1024 $  & (1,0) \\ 
			
			& Res $S_36$& Res $S_35$  & Res$S_37$& $1 \times 3 \times 3 \times 1024 $  & (1,0) \\  
			
			& Res $S_37$& Res $S_36$  & res $S_38$ & $1 \times 3 \times 3 \times 1024 $  & (1,0) \\ 
			
			& Res $S_38$& Res $S_37$  & Res $S_39$ & $1 \times 3 \times 3 \times 512 $  & (1,0) \\ 
			& Res $S_39$& Res $S_38$  & Deconv $S_31$ & $1 \times 3 \times 3 \times 256 $  & (2,1) \\

			& Deconv $S_31$& Res $S_39$  & Deconv $S_32$ & $1 \times 3 \times 3 \times 128 $  & (2,1) \\
			
			& Deconv $S_32$& Deconv $S_31$  & Deconv $S_33$ & $1 \times 3 \times 3 \times 64 $  & (2,1) \\

			& Deconv $S_33$& Deconv $S_32$  & Output $S_3$ & $1 \times 7 \times 7 \times 3 $  & (1,0) \\ \hline

		\end{tabular} 
	}
\end{table}

\subsection{Additional Experimental Results}
In this section, we present results reported in the main paper with a higher resolution. In addition, we present the following two additional baseline comparisons. 

\noindent \textbf{1. CycleGAN (Concat).} Input of multiple modalities are concatenated as channels. Operating on the concatenated input, cycleGAN is used to find the relevant transformation.  

\noindent \textbf{2. Image Fusion.} Input images are first fused using a wavelet-based image fusion technique.  In particular, wavelet coefficients of each input modality is found independently using db4 wavelet. In the wavelet domain, coefficients are fused by taking the average over all modalities. Fused coefficients are transformed back to the image domain by taking inverse wavelet transformation. Then, CycleGAN is operated on the fused image.

\begin{figure*}[!]
	\centering
	\begin{minipage}{.17\textwidth}
		\centering
		\includegraphics[width=3.0cm,height=3.0cm]{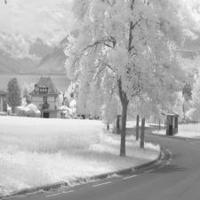}
		\captionsetup{labelformat=empty}
		\captionsetup{justification=centering}
		\caption*{\emph{Input(NIR)} \\ \quad}
	\end{minipage} 
	\begin{minipage}{.17\textwidth}
		\centering
		\includegraphics[width=3.0cm,height=3.0cm]{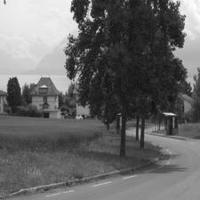}
		\captionsetup{labelformat=empty}
		\captionsetup{justification=centering}
		\caption*{{\emph{Input (Greyscale)}}\\ \quad}
	\end{minipage} 
	\begin{minipage}{.17\textwidth}
		\centering
		\includegraphics[width=3.0cm,height=3.0cm]{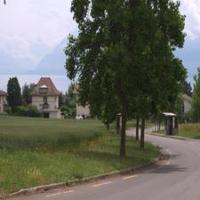}
		\captionsetup{labelformat=empty}
		\captionsetup{justification=centering}
		\caption*{{\emph{Target (RGB)}}\\ \quad}
	\end{minipage} 
	\begin{minipage}{.17\textwidth}
		\centering
		\includegraphics[width=3.0cm,height=3.0cm]{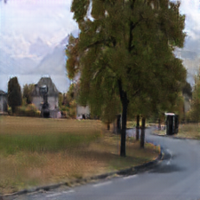}
		\captionsetup{labelformat=empty}
		\captionsetup{justification=centering}
		\caption*{{\emph{Ours}}\\ \quad}
	\end{minipage} 
	\begin{minipage}{.17\textwidth}
		\centering
		\includegraphics[width=3.0cm,height=3.0cm]{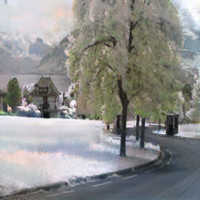}
		\captionsetup{labelformat=empty}
		\captionsetup{justification=centering}
		\caption*{{\emph{CycleGAN (NIR)}}\\ \quad}
	\end{minipage} 
	\begin{minipage}{.17\textwidth}
		\centering
		\includegraphics[width=3.0cm,height=3.0cm]{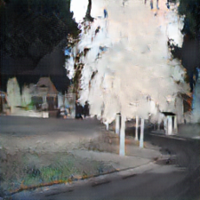}
		\captionsetup{labelformat=empty}
		\captionsetup{justification=centering}
		\caption*{{\emph{CycleGAN (Grey)} \cite{CycleGAN2017}}\\ \quad}
	\end{minipage} 
	\begin{minipage}{.17\textwidth}
		\centering
		\includegraphics[width=3.0cm,height=3.0cm]{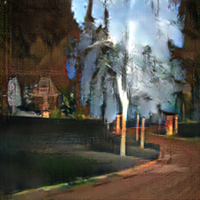}
		\captionsetup{labelformat=empty}
		\captionsetup{justification=centering}
		\caption*{{\emph{CycleGAN (Concat)} \cite{CycleGAN2017}}\\ \quad}
	\end{minipage} 
	\begin{minipage}{.17\textwidth}
		\centering
		\includegraphics[width=3.0cm,height=3.0cm]{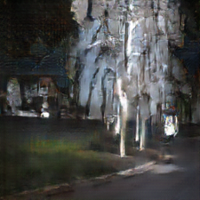}
		\captionsetup{labelformat=empty}
		\captionsetup{justification=centering}
		\caption*{{\emph{Image Fusion (CycleGAN)} \cite{CycleGAN2017}}\\ \quad}
	\end{minipage} 
	\begin{minipage}{.17\textwidth}
		\centering
		\includegraphics[width=3.0cm,height=3.0cm]{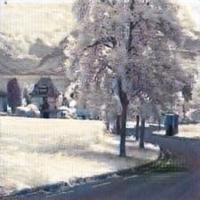}
		\captionsetup{labelformat=empty}
		\captionsetup{justification=centering}
		\caption*{{\emph{UNIT (NIR)} \\ \cite{NIPS2017_MING}}\\ \quad}
	\end{minipage} 
	\begin{minipage}{.17\textwidth}
		\centering
		\includegraphics[width=3.0cm,height=3.0cm]{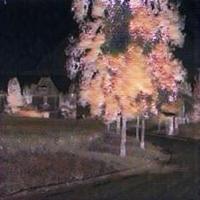}
		\captionsetup{labelformat=empty}
		\captionsetup{justification=centering}
		\caption*{{\emph{UNIT (Grey)}\\  \cite{NIPS2017_MING}}\\ \quad}
	\end{minipage}

	\vskip -8pt\caption{Results corresponding to the image colorization task (EPFL NIR-VIS Dataset - country category). Colorization achieved by the proposed method is closer to the ground truth compared with the baseline methods. This is most evident in the color of tree leaves, grass and the sky.  }\label{syn1}
\end{figure*}

\begin{figure*}[!]
	\centering
	\begin{minipage}{.17\textwidth}
		\centering
		\includegraphics[width=3.0cm,height=3.0cm]{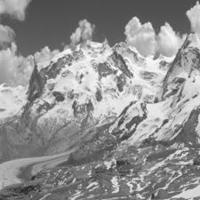}
		\captionsetup{labelformat=empty}
		\captionsetup{justification=centering}
		\caption*{\emph{Input(NIR)} \\ \quad}
	\end{minipage} 
	\begin{minipage}{.17\textwidth}
		\centering
		\includegraphics[width=3.0cm,height=3.0cm]{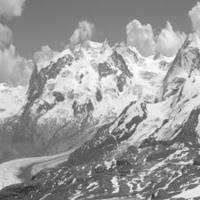}
		\captionsetup{labelformat=empty}
		\captionsetup{justification=centering}
		\caption*{{\emph{Input (Greyscale)}}\\ \quad}
	\end{minipage} 
	\begin{minipage}{.17\textwidth}
		\centering
		\includegraphics[width=3.0cm,height=3.0cm]{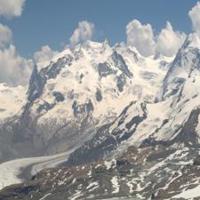}
		\captionsetup{labelformat=empty}
		\captionsetup{justification=centering}
		\caption*{{\emph{Target (RGB)}}\\ \quad}
	\end{minipage} 
	\begin{minipage}{.17\textwidth}
		\centering
		\includegraphics[width=3.0cm,height=3.0cm]{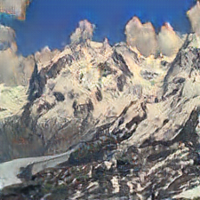}
		\captionsetup{labelformat=empty}
		\captionsetup{justification=centering}
		\caption*{{\emph{Ours}}\\ \quad}
	\end{minipage} 
	\begin{minipage}{.17\textwidth}
		\centering
		\includegraphics[width=3.0cm,height=3.0cm]{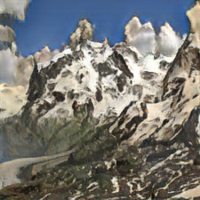}
		\captionsetup{labelformat=empty}
		\captionsetup{justification=centering}
		\caption*{{\emph{CycleGAN (NIR)}}\\ \quad}
	\end{minipage} 
	\begin{minipage}{.17\textwidth}
		\centering
		\includegraphics[width=3.0cm,height=3.0cm]{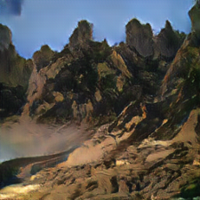}
		\captionsetup{labelformat=empty}
		\captionsetup{justification=centering}
		\caption*{{\emph{CycleGAN (Grey)} \cite{CycleGAN2017}}\\ \quad}
	\end{minipage} 
	\begin{minipage}{.17\textwidth}
		\centering
		\includegraphics[width=3.0cm,height=3.0cm]{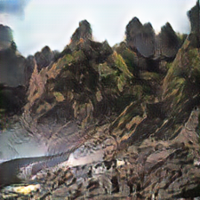}
		\captionsetup{labelformat=empty}
		\captionsetup{justification=centering}
		\caption*{{\emph{CycleGAN (Concat)} \cite{CycleGAN2017}}\\ \quad}
	\end{minipage} 
	\begin{minipage}{.17\textwidth}
		\centering
		\includegraphics[width=3.0cm,height=3.0cm]{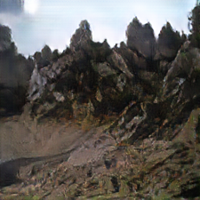}
		\captionsetup{labelformat=empty}
		\captionsetup{justification=centering}
		\caption*{{\emph{Image Fusion (CycleGAN)} \cite{CycleGAN2017}}\\ \quad}
	\end{minipage} 
	\begin{minipage}{.17\textwidth}
		\centering
		\includegraphics[width=3.0cm,height=3.0cm]{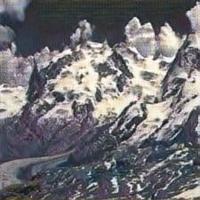}
		\captionsetup{labelformat=empty}
		\captionsetup{justification=centering}
		\caption*{{\emph{UNIT (NIR)} \\ \cite{NIPS2017_MING}}\\ \quad}
	\end{minipage} 
	\begin{minipage}{.17\textwidth}
		\centering
		\includegraphics[width=3.0cm,height=3.0cm]{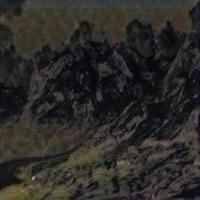}
		\captionsetup{labelformat=empty}
		\captionsetup{justification=centering}
		\caption*{{\emph{UNIT (Grey)}\\  \cite{NIPS2017_MING}}\\ \quad}
	\end{minipage}

	\vskip -8pt\caption{Results corresponding to the image colorization task (EPFL NIR-VIS Dataset - mountain category). Both CycleGAN(NIR) and UNIT(NIR) has produced results on par with the proposed method. However, valleys are colored in green in CycleGAN. UNIT has a dark gray shade imposed over the whole image. Comparatively the proposed method is more closer to the ground truth.  }\label{syn2}
\end{figure*}

\begin{figure*}[!]
	\centering
	\begin{minipage}{.17\textwidth}
		\centering
		\includegraphics[width=3.0cm,height=3.0cm]{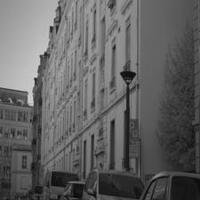}
		\captionsetup{labelformat=empty}
		\captionsetup{justification=centering}
		\caption*{\emph{Input(NIR)} \\ \quad}
	\end{minipage} 
	\begin{minipage}{.17\textwidth}
		\centering
		\includegraphics[width=3.0cm,height=3.0cm]{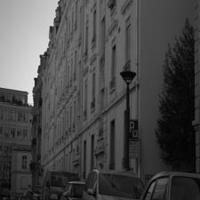}
		\captionsetup{labelformat=empty}
		\captionsetup{justification=centering}
		\caption*{{\emph{Input (Greyscale)}}\\ \quad}
	\end{minipage} 
	\begin{minipage}{.17\textwidth}
		\centering
		\includegraphics[width=3.0cm,height=3.0cm]{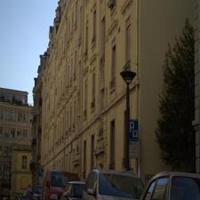}
		\captionsetup{labelformat=empty}
		\captionsetup{justification=centering}
		\caption*{{\emph{Target (RGB)}}\\ \quad}
	\end{minipage} 
	\begin{minipage}{.17\textwidth}
		\centering
		\includegraphics[width=3.0cm,height=3.0cm]{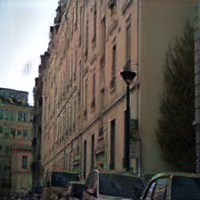}
		\captionsetup{labelformat=empty}
		\captionsetup{justification=centering}
		\caption*{{\emph{Ours}}\\ \quad}
	\end{minipage} 
	\begin{minipage}{.17\textwidth}
		\centering
		\includegraphics[width=3.0cm,height=3.0cm]{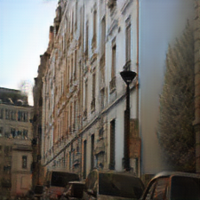}
		\captionsetup{labelformat=empty}
		\captionsetup{justification=centering}
		\caption*{{\emph{CycleGAN (NIR)}}\\ \quad}
	\end{minipage} 
	\begin{minipage}{.17\textwidth}
		\centering
		\includegraphics[width=3.0cm,height=3.0cm]{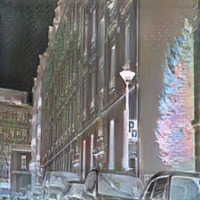}
		\captionsetup{labelformat=empty}
		\captionsetup{justification=centering}
		\caption*{{\emph{CycleGAN (Grey)} \cite{CycleGAN2017}}\\ \quad}
	\end{minipage} 
	\begin{minipage}{.17\textwidth}
		\centering
		\includegraphics[width=3.0cm,height=3.0cm]{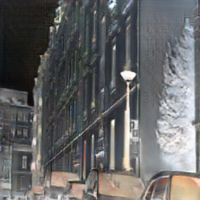}
		\captionsetup{labelformat=empty}
		\captionsetup{justification=centering}
		\caption*{{\emph{CycleGAN (Concat)} \cite{CycleGAN2017}}\\ \quad}
	\end{minipage} 
	\begin{minipage}{.17\textwidth}
		\centering
		\includegraphics[width=3.0cm,height=3.0cm]{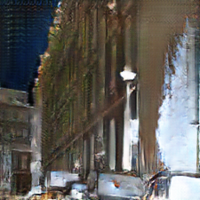}
		\captionsetup{labelformat=empty}
		\captionsetup{justification=centering}
		\caption*{{\emph{Image Fusion (CycleGAN)} \cite{CycleGAN2017}}\\ \quad}
	\end{minipage} 
	\begin{minipage}{.17\textwidth}
		\centering
		\includegraphics[width=3.0cm,height=3.0cm]{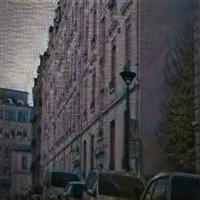}
		\captionsetup{labelformat=empty}
		\captionsetup{justification=centering}
		\caption*{{\emph{UNIT (NIR)} \\ \cite{NIPS2017_MING}}\\ \quad}
	\end{minipage} 
	\begin{minipage}{.17\textwidth}
		\centering
		\includegraphics[width=3.0cm,height=3.0cm]{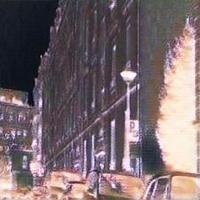}
		\captionsetup{labelformat=empty}
		\captionsetup{justification=centering}
		\caption*{{\emph{UNIT (Grey)}\\  \cite{NIPS2017_MING}}\\ \quad}
	\end{minipage}

	\vskip -8pt\caption{Results corresponding to the image colorization task (EPFL NIR-VIS Dataset - urban category). Color of sky, tree and the building of the proposed method's output is more closer to the ground truth compared to other baselines. }\label{syn3}
\end{figure*}

\begin{figure*}[!]
	\centering
	\begin{minipage}{.17\textwidth}
		\centering
		\includegraphics[width=3.0cm,height=3.0cm]{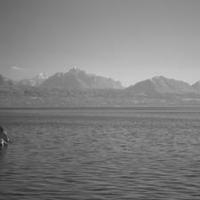}
		\captionsetup{labelformat=empty}
		\captionsetup{justification=centering}
		\caption*{\emph{Input(NIR)} \\ \quad}
	\end{minipage} 
	\begin{minipage}{.17\textwidth}
		\centering
		\includegraphics[width=3.0cm,height=3.0cm]{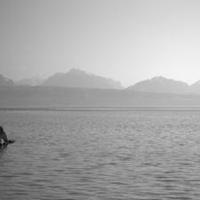}
		\captionsetup{labelformat=empty}
		\captionsetup{justification=centering}
		\caption*{{\emph{Input (Greyscale)}}\\ \quad}
	\end{minipage} 
	\begin{minipage}{.17\textwidth}
		\centering
		\includegraphics[width=3.0cm,height=3.0cm]{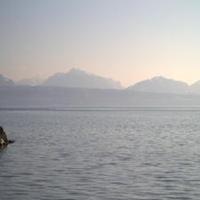}
		\captionsetup{labelformat=empty}
		\captionsetup{justification=centering}
		\caption*{{\emph{Target (RGB)}}\\ \quad}
	\end{minipage} 
	\begin{minipage}{.17\textwidth}
		\centering
		\includegraphics[width=3.0cm,height=3.0cm]{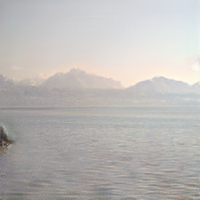}
		\captionsetup{labelformat=empty}
		\captionsetup{justification=centering}
		\caption*{{\emph{Ours}}\\ \quad}
	\end{minipage} 
	\begin{minipage}{.17\textwidth}
		\centering
		\includegraphics[width=3.0cm,height=3.0cm]{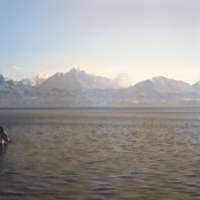}
		\captionsetup{labelformat=empty}
		\captionsetup{justification=centering}
		\caption*{{\emph{CycleGAN (NIR)}}\\ \quad}
	\end{minipage} 
	\begin{minipage}{.17\textwidth}
		\centering
		\includegraphics[width=3.0cm,height=3.0cm]{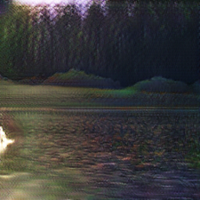}
		\captionsetup{labelformat=empty}
		\captionsetup{justification=centering}
		\caption*{{\emph{CycleGAN (Grey)} \cite{CycleGAN2017}}\\ \quad}
	\end{minipage} 
	\begin{minipage}{.17\textwidth}
		\centering
		\includegraphics[width=3.0cm,height=3.0cm]{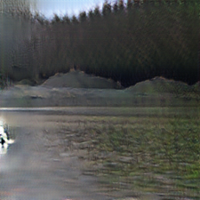}
		\captionsetup{labelformat=empty}
		\captionsetup{justification=centering}
		\caption*{{\emph{CycleGAN (Concat)} \cite{CycleGAN2017}}\\ \quad}
	\end{minipage} 
	\begin{minipage}{.17\textwidth}
		\centering
		\includegraphics[width=3.0cm,height=3.0cm]{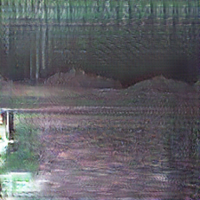}
		\captionsetup{labelformat=empty}
		\captionsetup{justification=centering}
		\caption*{{\emph{Image Fusion (CycleGAN)} \cite{CycleGAN2017}}\\ \quad}
	\end{minipage} 
	\begin{minipage}{.17\textwidth}
		\centering
		\includegraphics[width=3.0cm,height=3.0cm]{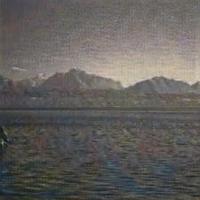}
		\captionsetup{labelformat=empty}
		\captionsetup{justification=centering}
		\caption*{{\emph{UNIT (NIR)} \\ \cite{NIPS2017_MING}}\\ \quad}
	\end{minipage} 
	\begin{minipage}{.17\textwidth}
		\centering
		\includegraphics[width=3.0cm,height=3.0cm]{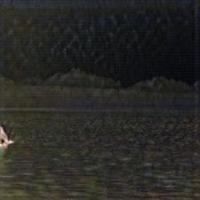}
		\captionsetup{labelformat=empty}
		\captionsetup{justification=centering}
		\caption*{{\emph{UNIT (Grey)}\\  \cite{NIPS2017_MING}}\\ \quad}
	\end{minipage}

	\vskip -8pt\caption{Results corresponding to the image colorization task (EPFL NIR-VIS Dataset - water category). CycleGAN (NIR) has the closest result in terms of quality to the proposed method. However, water has a yellow shade mixed to it and the sky seems more blue compared to the ground truth. These differences are not present in our output. }\label{syn4}
\end{figure*}

\begin{figure*}[!]
	\centering
	
	\begin{minipage}{.17\textwidth}
		\centering
		\includegraphics[width=3.0cm,height=3.0cm]{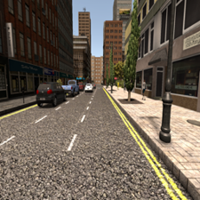}
		\captionsetup{labelformat=empty}
		\captionsetup{justification=centering}
		\caption*{\emph{Input(RGB)} \\ \quad}
	\end{minipage} 
	\begin{minipage}{.17\textwidth}
		\centering
		\includegraphics[width=3.0cm,height=3.0cm]{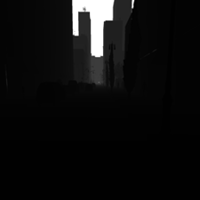}
		\captionsetup{labelformat=empty}
		\captionsetup{justification=centering}
		\caption*{{\emph{Input (Depth)}}\\ \quad}
	\end{minipage} 
	\begin{minipage}{.17\textwidth}
		\centering
		\includegraphics[width=3.0cm,height=3.0cm]{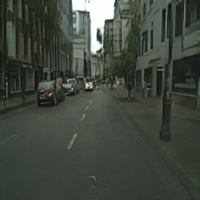}
		\captionsetup{labelformat=empty}
		\captionsetup{justification=centering}
		\caption*{{\emph{Ours}}\\ \quad}
	\end{minipage} 
	\begin{minipage}{.17\textwidth}
		\centering
		\includegraphics[width=3.0cm,height=3.0cm]{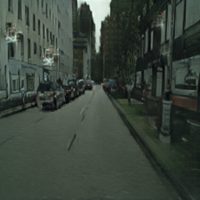}
		\captionsetup{labelformat=empty}
		\captionsetup{justification=centering}
		\caption*{{\emph{CycleGAN (RGB)}}\\ \quad}
	\end{minipage}

	\begin{minipage}{.17\textwidth}
		\centering
		\includegraphics[width=3.0cm,height=3.0cm]{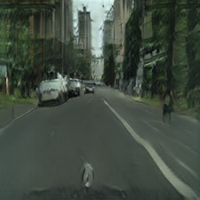}
		\captionsetup{labelformat=empty}
		\captionsetup{justification=centering}
		\caption*{{\emph{CycleGAN (Depth)} \cite{CycleGAN2017}}\\ \quad}
	\end{minipage} 
	\begin{minipage}{.17\textwidth}
		\centering
		\includegraphics[width=3.0cm,height=3.0cm]{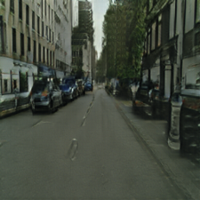}
		\captionsetup{labelformat=empty}
		\captionsetup{justification=centering}
		\caption*{{\emph{CycleGAN (Concat)} \cite{CycleGAN2017}}\\ \quad}
	\end{minipage} 
	\begin{minipage}{.17\textwidth}
		\centering
		\includegraphics[width=3.0cm,height=3.0cm]{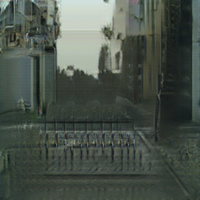}
		\captionsetup{labelformat=empty}
		\captionsetup{justification=centering}
		\caption*{{\emph{Image Fusion (CycleGAN)} \cite{CycleGAN2017}}\\ \quad}
	\end{minipage} 
	\begin{minipage}{.17\textwidth}
		\centering
		\includegraphics[width=3.0cm,height=3.0cm]{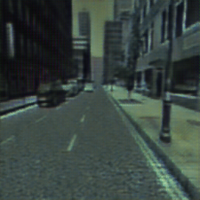}
		\captionsetup{labelformat=empty}
		\captionsetup{justification=centering}
		\caption*{{\emph{UNIT (RGB)} \\ \cite{NIPS2017_MING}}\\ \quad}
	\end{minipage} 
	\begin{minipage}{.17\textwidth}
		\centering
		\includegraphics[width=3.0cm,height=3.0cm]{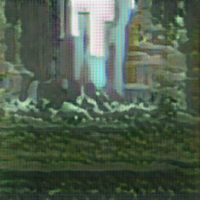}
		\captionsetup{labelformat=empty}
		\captionsetup{justification=centering}
		\caption*{{\emph{UNIT (Depth)}\\  \cite{NIPS2017_MING}}\\ \quad}
	\end{minipage}

	\vskip -8pt\caption{Results corresponding to the Synthetic-to-Real translation task (sample 00400). CycleGAN (RGB) and CycleGAN (Concat) have distorted constructions. The pole color is inverted; distortions are visible on cars and buildings; building far away are mistaken as trees. UNIT(RGB) has only changed the color scheme of the output. Characteristics of artificial graphics are apparent (texture of the road for an example). Proposed method produces a more realistic construction comparatively.   }\label{syn5}
\end{figure*}

\begin{figure*}[!]
	\centering
	
	\begin{minipage}{.17\textwidth}
		\centering
		\includegraphics[width=3.0cm,height=3.0cm]{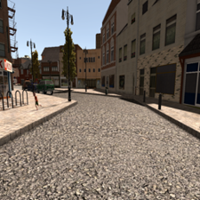}
		\captionsetup{labelformat=empty}
		\captionsetup{justification=centering}
		\caption*{\emph{Input(RGB)} \\ \quad}
	\end{minipage} 
	\begin{minipage}{.17\textwidth}
		\centering
		\includegraphics[width=3.0cm,height=3.0cm]{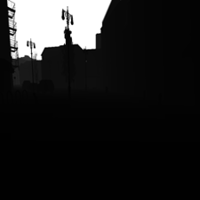}
		\captionsetup{labelformat=empty}
		\captionsetup{justification=centering}
		\caption*{{\emph{Input (Depth)}}\\ \quad}
	\end{minipage} 
	\begin{minipage}{.17\textwidth}
		\centering
		\includegraphics[width=3.0cm,height=3.0cm]{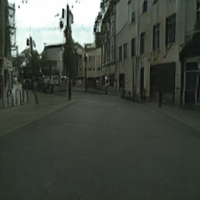}
		\captionsetup{labelformat=empty}
		\captionsetup{justification=centering}
		\caption*{{\emph{Ours}}\\ \quad}
	\end{minipage} 
	\begin{minipage}{.17\textwidth}
		\centering
		\includegraphics[width=3.0cm,height=3.0cm]{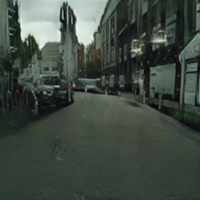}
		\captionsetup{labelformat=empty}
		\captionsetup{justification=centering}
		\caption*{{\emph{CycleGAN (RGB)}}\\ \quad}
	\end{minipage}

	\begin{minipage}{.17\textwidth}
		\centering
		\includegraphics[width=3.0cm,height=3.0cm]{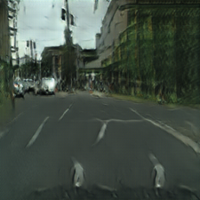}
		\captionsetup{labelformat=empty}
		\captionsetup{justification=centering}
		\caption*{{\emph{CycleGAN (Depth)} \cite{CycleGAN2017}}\\ \quad}
	\end{minipage} 
	\begin{minipage}{.17\textwidth}
		\centering
		\includegraphics[width=3.0cm,height=3.0cm]{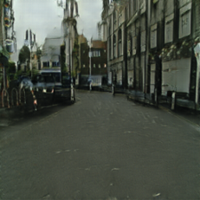}
		\captionsetup{labelformat=empty}
		\captionsetup{justification=centering}
		\caption*{{\emph{CycleGAN (Concat)} \cite{CycleGAN2017}}\\ \quad}
	\end{minipage} 
	\begin{minipage}{.17\textwidth}
		\centering
		\includegraphics[width=3.0cm,height=3.0cm]{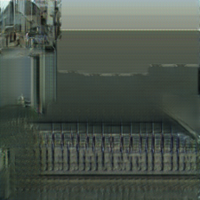}
		\captionsetup{labelformat=empty}
		\captionsetup{justification=centering}
		\caption*{{\emph{Image Fusion (CycleGAN)} \cite{CycleGAN2017}}\\ \quad}
	\end{minipage} 
	\begin{minipage}{.17\textwidth}
		\centering
		\includegraphics[width=3.0cm,height=3.0cm]{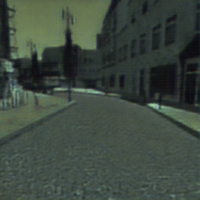}
		\captionsetup{labelformat=empty}
		\captionsetup{justification=centering}
		\caption*{{\emph{UNIT (RGB)} \\ \cite{NIPS2017_MING}}\\ \quad}
	\end{minipage} 
	\begin{minipage}{.17\textwidth}
		\centering
		\includegraphics[width=3.0cm,height=3.0cm]{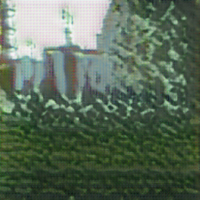}
		\captionsetup{labelformat=empty}
		\captionsetup{justification=centering}
		\caption*{{\emph{UNIT (Depth)}\\  \cite{NIPS2017_MING}}\\ \quad}
	\end{minipage}

	\vskip -8pt\caption{Results corresponding to the Synthetic-to-Real translation task (sample 00600). In addition to points made in Figure~\ref{syn5}, note that both CycleGAN (RGB) and CycleGAN (Concat)  have erroneously generated clutter in the left pavement. }\label{syn6}
\end{figure*}

\begin{figure*}[!]
	\centering
	
	\begin{minipage}{.17\textwidth}
		\centering
		\includegraphics[width=3.0cm,height=3.0cm]{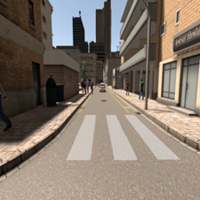}
		\captionsetup{labelformat=empty}
		\captionsetup{justification=centering}
		\caption*{\emph{Input(RGB)} \\ \quad}
	\end{minipage} 
	\begin{minipage}{.17\textwidth}
		\centering
		\includegraphics[width=3.0cm,height=3.0cm]{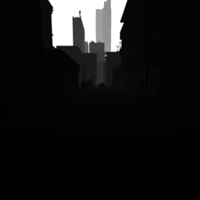}
		\captionsetup{labelformat=empty}
		\captionsetup{justification=centering}
		\caption*{{\emph{Input (Depth)}}\\ \quad}
	\end{minipage} 
	\begin{minipage}{.17\textwidth}
		\centering
		\includegraphics[width=3.0cm,height=3.0cm]{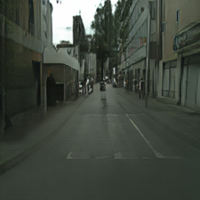}
		\captionsetup{labelformat=empty}
		\captionsetup{justification=centering}
		\caption*{{\emph{Ours}}\\ \quad}
	\end{minipage} 
	\begin{minipage}{.17\textwidth}
		\centering
		\includegraphics[width=3.0cm,height=3.0cm]{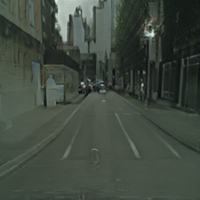}
		\captionsetup{labelformat=empty}
		\captionsetup{justification=centering}
		\caption*{{\emph{CycleGAN (RGB)}}\\ \quad}
	\end{minipage}

	\begin{minipage}{.17\textwidth}
		\centering
		\includegraphics[width=3.0cm,height=3.0cm]{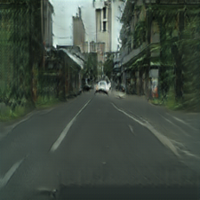}
		\captionsetup{labelformat=empty}
		\captionsetup{justification=centering}
		\caption*{{\emph{CycleGAN (Depth)} \cite{CycleGAN2017}}\\ \quad}
	\end{minipage} 
	\begin{minipage}{.17\textwidth}
		\centering
		\includegraphics[width=3.0cm,height=3.0cm]{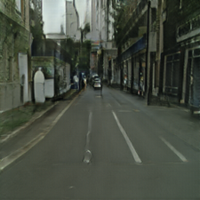}
		\captionsetup{labelformat=empty}
		\captionsetup{justification=centering}
		\caption*{{\emph{CycleGAN (Concat)} \cite{CycleGAN2017}}\\ \quad}
	\end{minipage} 
	\begin{minipage}{.17\textwidth}
		\centering
		\includegraphics[width=3.0cm,height=3.0cm]{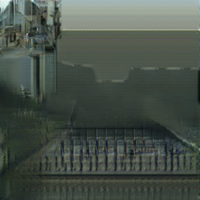}
		\captionsetup{labelformat=empty}
		\captionsetup{justification=centering}
		\caption*{{\emph{Image Fusion (CycleGAN)} \cite{CycleGAN2017}}\\ \quad}
	\end{minipage} 
	\begin{minipage}{.17\textwidth}
		\centering
		\includegraphics[width=3.0cm,height=3.0cm]{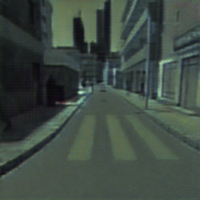}
		\captionsetup{labelformat=empty}
		\captionsetup{justification=centering}
		\caption*{{\emph{UNIT (RGB)} \\ \cite{NIPS2017_MING}}\\ \quad}
	\end{minipage} 
	\begin{minipage}{.17\textwidth}
		\centering
		\includegraphics[width=3.0cm,height=3.0cm]{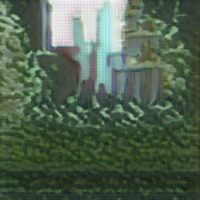}
		\captionsetup{labelformat=empty}
		\captionsetup{justification=centering}
		\caption*{{\emph{UNIT (Depth)}\\  \cite{NIPS2017_MING}}\\ \quad}
	\end{minipage}

	\vskip -8pt\caption{Results corresponding to the Synthetic-to-Real translation task (sample 00800). CycleGAN (RGB) output is similar to the output of the proposed method to an extent in this case. Note there are some distortions present in walls in either side of the street in CycleGAN (RGB) output image.   }\label{syn7}
\end{figure*}

\begin{figure*}[!]
	\centering
	
	\begin{minipage}{.17\textwidth}
		\centering
		\includegraphics[width=3.0cm,height=3.0cm]{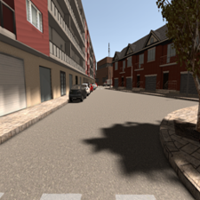}
		\captionsetup{labelformat=empty}
		\captionsetup{justification=centering}
		\caption*{\emph{Input(RGB)} \\ \quad}
	\end{minipage} 
	\begin{minipage}{.17\textwidth}
		\centering
		\includegraphics[width=3.0cm,height=3.0cm]{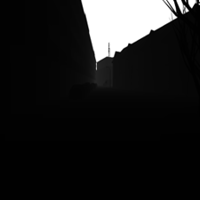}
		\captionsetup{labelformat=empty}
		\captionsetup{justification=centering}
		\caption*{{\emph{Input (Depth)}}\\ \quad}
	\end{minipage} 
	\begin{minipage}{.17\textwidth}
		\centering
		\includegraphics[width=3.0cm,height=3.0cm]{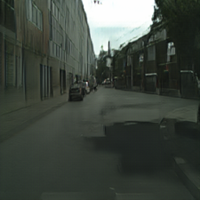}
		\captionsetup{labelformat=empty}
		\captionsetup{justification=centering}
		\caption*{{\emph{Ours}}\\ \quad}
	\end{minipage} 
	\begin{minipage}{.17\textwidth}
		\centering
		\includegraphics[width=3.0cm,height=3.0cm]{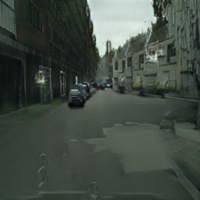}
		\captionsetup{labelformat=empty}
		\captionsetup{justification=centering}
		\caption*{{\emph{CycleGAN (RGB)}}\\ \quad}
	\end{minipage}

	\begin{minipage}{.17\textwidth}
		\centering
		\includegraphics[width=3.0cm,height=3.0cm]{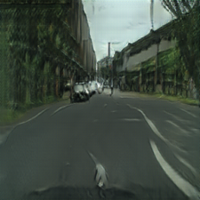}
		\captionsetup{labelformat=empty}
		\captionsetup{justification=centering}
		\caption*{{\emph{CycleGAN (Depth)} \cite{CycleGAN2017}}\\ \quad}
	\end{minipage} 
	\begin{minipage}{.17\textwidth}
		\centering
		\includegraphics[width=3.0cm,height=3.0cm]{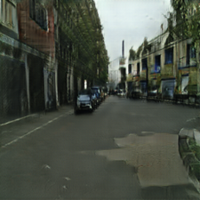}
		\captionsetup{labelformat=empty}
		\captionsetup{justification=centering}
		\caption*{{\emph{CycleGAN (Concat)} \cite{CycleGAN2017}}\\ \quad}
	\end{minipage} 
	\begin{minipage}{.17\textwidth}
		\centering
		\includegraphics[width=3.0cm,height=3.0cm]{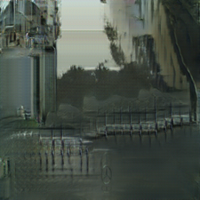}
		\captionsetup{labelformat=empty}
		\captionsetup{justification=centering}
		\caption*{{\emph{Image Fusion (CycleGAN)} \cite{CycleGAN2017}}\\ \quad}
	\end{minipage} 
	\begin{minipage}{.17\textwidth}
		\centering
		\includegraphics[width=3.0cm,height=3.0cm]{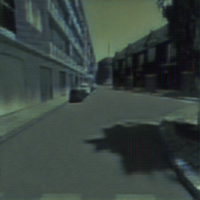}
		\captionsetup{labelformat=empty}
		\captionsetup{justification=centering}
		\caption*{{\emph{UNIT (RGB)} \\ \cite{NIPS2017_MING}}\\ \quad}
	\end{minipage} 
	\begin{minipage}{.17\textwidth}
		\centering
		\includegraphics[width=3.0cm,height=3.0cm]{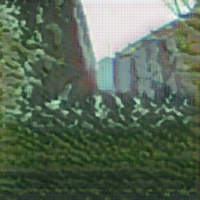}
		\captionsetup{labelformat=empty}
		\captionsetup{justification=centering}
		\caption*{{\emph{UNIT (Depth)}\\  \cite{NIPS2017_MING}}\\ \quad}
	\end{minipage}

	\vskip -8pt\caption{Results corresponding to the Synthetic-to-Real translation task (sample 00700). Both CycleGAN (RGB) and CycleGAN (Concat) have failed to identify the shadow of the tree in their respective outputs. Compared to the output of the proposed method, both these outputs have considerable distortions on the walls at either side of the street.   }\label{syn8}
\end{figure*}

\begin{figure*}[!]
	\centering
	\begin{minipage}{.17\textwidth}
		\centering
		\includegraphics[width=3.0cm,height=3.0cm]{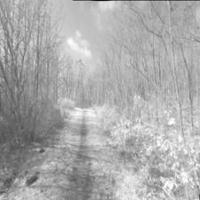}
		\captionsetup{labelformat=empty}
		\captionsetup{justification=centering}
		\caption*{\emph{Input(NIR)} \\ \quad}
	\end{minipage} 
	\begin{minipage}{.17\textwidth}
		\centering
		\includegraphics[width=3.0cm,height=3.0cm]{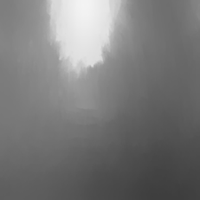}
		\captionsetup{labelformat=empty}
		\captionsetup{justification=centering}
		\caption*{{\emph{Input (Depth)}}\\ \quad}
	\end{minipage} 
	\begin{minipage}{.17\textwidth}
		\centering
		\includegraphics[width=3.0cm,height=3.0cm]{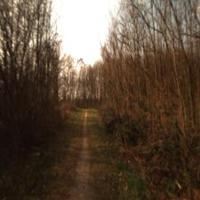}
		\captionsetup{labelformat=empty}
		\captionsetup{justification=centering}
		\caption*{{\emph{Target (RGB)}}\\ \quad}
	\end{minipage} 
	\begin{minipage}{.17\textwidth}
		\centering
		\includegraphics[width=3.0cm,height=3.0cm]{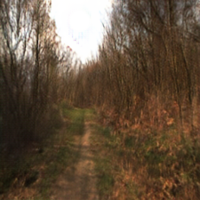}
		\captionsetup{labelformat=empty}
		\captionsetup{justification=centering}
		\caption*{{\emph{Ours}}\\ \quad}
	\end{minipage} 
	\begin{minipage}{.17\textwidth}
		\centering
		\includegraphics[width=3.0cm,height=3.0cm]{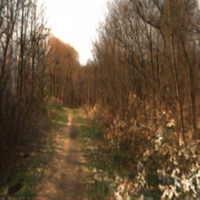}
		\captionsetup{labelformat=empty}
		\captionsetup{justification=centering}
		\caption*{{\emph{CycleGAN (NIR)}}\\ \quad}
	\end{minipage} 
	\begin{minipage}{.17\textwidth}
		\centering
		\includegraphics[width=3.0cm,height=3.0cm]{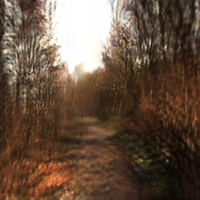}
		\captionsetup{labelformat=empty}
		\captionsetup{justification=centering}
		\caption*{{\emph{CycleGAN (Depth)} \cite{CycleGAN2017}}\\ \quad}
	\end{minipage} 
	\begin{minipage}{.17\textwidth}
		\centering
		\includegraphics[width=3.0cm,height=3.0cm]{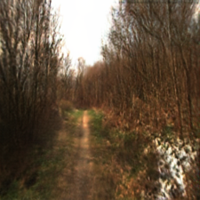}
		\captionsetup{labelformat=empty}
		\captionsetup{justification=centering}
		\caption*{{\emph{CycleGAN (Concat)} \cite{CycleGAN2017}}\\ \quad}
	\end{minipage} 
	\begin{minipage}{.17\textwidth}
		\centering
		\includegraphics[width=3.0cm,height=3.0cm]{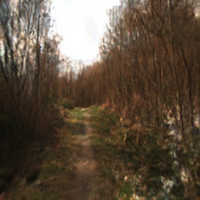}
		\captionsetup{labelformat=empty}
		\captionsetup{justification=centering}
		\caption*{{\emph{Image Fusion (CycleGAN)} \cite{CycleGAN2017}}\\ \quad}
	\end{minipage} 
	\begin{minipage}{.17\textwidth}
		\centering
		\includegraphics[width=3.0cm,height=3.0cm]{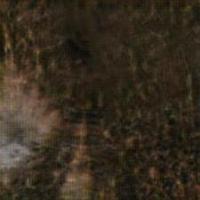}
		\captionsetup{labelformat=empty}
		\captionsetup{justification=centering}
		\caption*{{\emph{UNIT (NIR)} \\ \cite{NIPS2017_MING}}\\ \quad}
	\end{minipage} 
	\begin{minipage}{.17\textwidth}
		\centering
		\includegraphics[width=3.0cm,height=3.0cm]{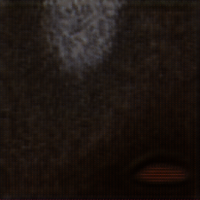}
		\captionsetup{labelformat=empty}
		\captionsetup{justification=centering}
		\caption*{{\emph{UNIT (Depth)}\\  \cite{NIPS2017_MING}}\\ \quad}
	\end{minipage}

	\vskip -8pt\caption{Results corresponding to the hyperspectral-to-visible translation task (image b118-047). White color artifacts present in CycleGAN (NIR) is more apparent in CycleGAN (Concat). Image Fusion (CycleGAN) has produced an image with less artifacts. However, the proposed method produces the closest reconstruction to the ground truth.  }\label{syn9}
\end{figure*}

\begin{figure*}[!]
	\centering
	\begin{minipage}{.17\textwidth}
		\centering
		\includegraphics[width=3.0cm,height=3.0cm]{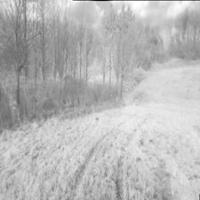}
		\captionsetup{labelformat=empty}
		\captionsetup{justification=centering}
		\caption*{\emph{Input(NIR)} \\ \quad}
	\end{minipage} 
	\begin{minipage}{.17\textwidth}
		\centering
		\includegraphics[width=3.0cm,height=3.0cm]{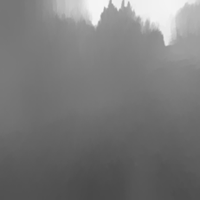}
		\captionsetup{labelformat=empty}
		\captionsetup{justification=centering}
		\caption*{{\emph{Input (Depth)}}\\ \quad}
	\end{minipage} 
	\begin{minipage}{.17\textwidth}
		\centering
		\includegraphics[width=3.0cm,height=3.0cm]{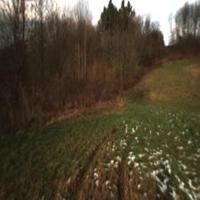}
		\captionsetup{labelformat=empty}
		\captionsetup{justification=centering}
		\caption*{{\emph{Target (RGB)}}\\ \quad}
	\end{minipage} 
	\begin{minipage}{.17\textwidth}
		\centering
		\includegraphics[width=3.0cm,height=3.0cm]{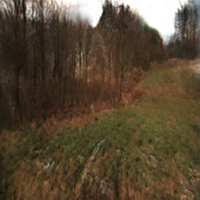}
		\captionsetup{labelformat=empty}
		\captionsetup{justification=centering}
		\caption*{{\emph{Ours}}\\ \quad}
	\end{minipage} 
	\begin{minipage}{.17\textwidth}
		\centering
		\includegraphics[width=3.0cm,height=3.0cm]{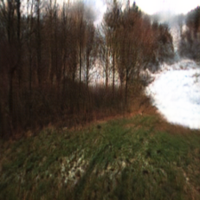}
		\captionsetup{labelformat=empty}
		\captionsetup{justification=centering}
		\caption*{{\emph{CycleGAN (NIR)}}\\ \quad}
	\end{minipage} 
	\begin{minipage}{.17\textwidth}
		\centering
		\includegraphics[width=3.0cm,height=3.0cm]{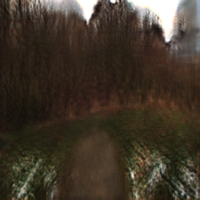}
		\captionsetup{labelformat=empty}
		\captionsetup{justification=centering}
		\caption*{{\emph{CycleGAN (Depth)} \cite{CycleGAN2017}}\\ \quad}
	\end{minipage} 
	\begin{minipage}{.17\textwidth}
		\centering
		\includegraphics[width=3.0cm,height=3.0cm]{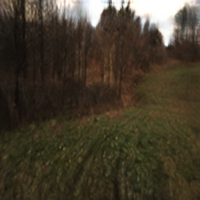}
		\captionsetup{labelformat=empty}
		\captionsetup{justification=centering}
		\caption*{{\emph{CycleGAN (Concat)} \cite{CycleGAN2017}}\\ \quad}
	\end{minipage} 
	\begin{minipage}{.17\textwidth}
		\centering
		\includegraphics[width=3.0cm,height=3.0cm]{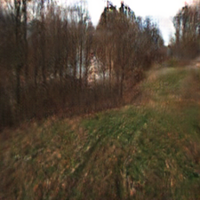}
		\captionsetup{labelformat=empty}
		\captionsetup{justification=centering}
		\caption*{{\emph{Image Fusion (CycleGAN)} \cite{CycleGAN2017}}\\ \quad}
	\end{minipage} 
	\begin{minipage}{.17\textwidth}
		\centering
		\includegraphics[width=3.0cm,height=3.0cm]{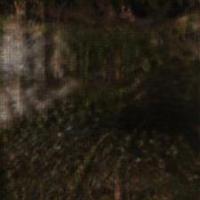}
		\captionsetup{labelformat=empty}
		\captionsetup{justification=centering}
		\caption*{{\emph{UNIT (NIR)} \\ \cite{NIPS2017_MING}}\\ \quad}
	\end{minipage} 
	\begin{minipage}{.17\textwidth}
		\centering
		\includegraphics[width=3.0cm,height=3.0cm]{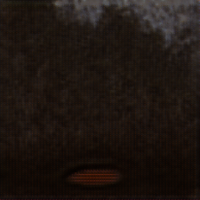}
		\captionsetup{labelformat=empty}
		\captionsetup{justification=centering}
		\caption*{{\emph{UNIT (Depth)}\\  \cite{NIPS2017_MING}}\\ \quad}
	\end{minipage}

	\vskip -8pt\caption{Results corresponding to the hyperspectral-to-visible translation task (image b197-61). CycleGAN (Concat) have produced a result on par with the proposed method while there are observable deficiencies in other outputs. However, in terms of colors (trees at the left and the grass), latter is more closer to the ground truth. }\label{syn10}
\end{figure*}

\begin{figure*}[!]
	\centering
	\begin{minipage}{.17\textwidth}
		\centering
		\includegraphics[width=3.0cm,height=3.0cm]{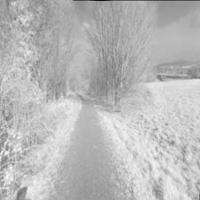}
		\captionsetup{labelformat=empty}
		\captionsetup{justification=centering}
		\caption*{\emph{Input(NIR)} \\ \quad}
	\end{minipage} 
	\begin{minipage}{.17\textwidth}
		\centering
		\includegraphics[width=3.0cm,height=3.0cm]{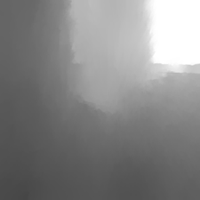}
		\captionsetup{labelformat=empty}
		\captionsetup{justification=centering}
		\caption*{{\emph{Input (Depth)}}\\ \quad}
	\end{minipage} 
	\begin{minipage}{.17\textwidth}
		\centering
		\includegraphics[width=3.0cm,height=3.0cm]{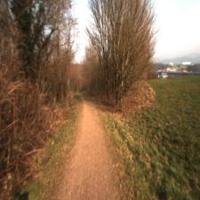}
		\captionsetup{labelformat=empty}
		\captionsetup{justification=centering}
		\caption*{{\emph{Target (RGB)}}\\ \quad}
	\end{minipage} 
	\begin{minipage}{.17\textwidth}
		\centering
		\includegraphics[width=3.0cm,height=3.0cm]{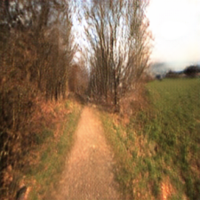}
		\captionsetup{labelformat=empty}
		\captionsetup{justification=centering}
		\caption*{{\emph{Ours}}\\ \quad}
	\end{minipage} 
	\begin{minipage}{.17\textwidth}
		\centering
		\includegraphics[width=3.0cm,height=3.0cm]{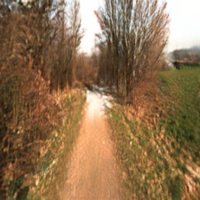}
		\captionsetup{labelformat=empty}
		\captionsetup{justification=centering}
		\caption*{{\emph{CycleGAN (NIR)}}\\ \quad}
	\end{minipage} 
	\begin{minipage}{.17\textwidth}
		\centering
		\includegraphics[width=3.0cm,height=3.0cm]{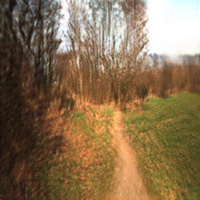}
		\captionsetup{labelformat=empty}
		\captionsetup{justification=centering}
		\caption*{{\emph{CycleGAN (Depth)} \cite{CycleGAN2017}}\\ \quad}
	\end{minipage} 
	\begin{minipage}{.17\textwidth}
		\centering
		\includegraphics[width=3.0cm,height=3.0cm]{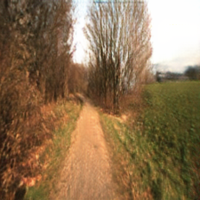}
		\captionsetup{labelformat=empty}
		\captionsetup{justification=centering}
		\caption*{{\emph{CycleGAN (Concat)} \cite{CycleGAN2017}}\\ \quad}
	\end{minipage} 
	\begin{minipage}{.17\textwidth}
		\centering
		\includegraphics[width=3.0cm,height=3.0cm]{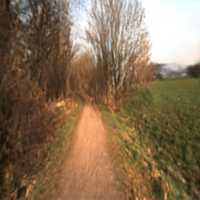}
		\captionsetup{labelformat=empty}
		\captionsetup{justification=centering}
		\caption*{{\emph{Image Fusion (CycleGAN)} \cite{CycleGAN2017}}\\ \quad}
	\end{minipage} 
	\begin{minipage}{.17\textwidth}
		\centering
		\includegraphics[width=3.0cm,height=3.0cm]{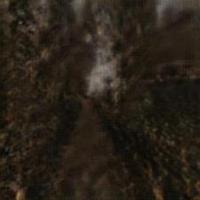}
		\captionsetup{labelformat=empty}
		\captionsetup{justification=centering}
		\caption*{{\emph{UNIT (NIR)} \\ \cite{NIPS2017_MING}}\\ \quad}
	\end{minipage} 
	\begin{minipage}{.17\textwidth}
		\centering
		\includegraphics[width=3.0cm,height=3.0cm]{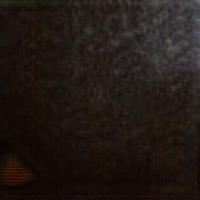}
		\captionsetup{labelformat=empty}
		\captionsetup{justification=centering}
		\caption*{{\emph{UNIT (Depth)}\\  \cite{NIPS2017_MING}}\\ \quad}
	\end{minipage}

	\vskip -8pt\caption{Results corresponding to the hyperspectral-to-visible translation task (image b304-36).} Both CycleGAN (Concat) and the proposed method produces results of similar quality for this image. Therefore, feature fusion and pixel-level fusion both have similar performances for this image.\label{syn11}
\end{figure*}

Results corresponding to the image colorization task are shown in Figures~\ref{syn1},\ref{syn2},\ref{syn3} and \ref{syn4}. In all these cases, the proposed method yields more realistic colorization. In Figures \ref{syn5},\ref{syn6},\ref{syn7} and \ref{syn8} results obtained for synthetic-to-real translation are shown. As described in the main paper, the proposed method has performed a translation of higher quality in this task as well. The third task, hyper-spectral-to-visual image translation, is the easier task among the three tasks. Therefore, CycleGAN (NIR), CycleGAN(Concat) and Image Fusion(CycleGAN) are able to produce results on par with the proposed method (Figure~\ref{syn11}). However, in images \ref{syn9} and \ref{syn10}, the proposed method is able to produce images of distinguishable higher quality compared with the baselines. 

\subsection{Pixel-level Fusion vs Feature Fusion} \label{sec:fvs}
In the proposed network, information of input modalities are fused at the beginning of the encoder sub-network. In principle, fusion can carried out as  pixel-level fusion, feature fusion or decision fusion \cite{Mitchell_Book}. Since the task in hand takes the form of image reconstruction, decision fusion is not applicable. In our method, we utilize the feature fusion technique where we first extract some feature maps from each modality and fuse them together using a convolution operation. In principle, it is also possible to use pixel-level fusion for this task.
For example, pixel-level fusion can be performed by training a CycleGAN model where a concatenation of all $n$ inputs are provided as the input to the network (\textit{CycleGAN (Concat)} in our experiments). 

However, when the input modalities are from incompatible domains, pixel-level fusion may result in incoherent reconstructions. In order to illustrate this, we direct readers' attention to \textit{CycleGAN (Concat)} results shown in Figures \ref{syn1} to \ref{syn11}. Compared to the results of CycleGAN in Figures~\ref{syn1},\ref{syn2},\ref{syn3},\ref{syn4} the colorization in cycleGAN-fused is less realistic. For example, CycleGAN operating on the NIR images has captured the colorization of the mountain image in Figure~\ref{syn2} much better than the fused version has. In this case, pixel-level fusion in fact has deteriorated the performance as compared to the original case. This trend can be observed across all three tasks. However, the performance of CycleGAN(Concat) is reasonable in most images (except for Figure~\ref{syn9}), in the hyper-spectral-to-visible translation task. In this special case, pixel-level fusion has worked effectively.

\section*{Acknowledgement}
This work was supported by US Office of Naval Research (ONR) Grant
YIP N00014-16-1-3134.

{\small
\bibliographystyle{ieee}

\bibliography{egbib}
}

\end{document}